\documentclass{eusflat2021}
\usepackage{cite}
\usepackage{amsmath,amssymb,amsfonts,lipsum,float,gensymb,cancel,mathrsfs,url,cancel,amsfonts,textcomp,xcolor,wrapfig,sidecap,pdfpages,fancyhdr,booktabs,multirow,capt-of,tabularx,changepage,transparent,etoolbox}
\usepackage{algorithmic}
\usepackage{graphicx}
\usepackage{textcomp}
\usepackage[export]{adjustbox}

\title{\bf DronePose: The identification, segmentation, and orientation detection of drones via neural networks}
\author{Stirling Scholes$^1$, Alice Ruget$^1$, Germ\'an Mora-Mart\'in$^2$, Feng Zhu$^1$, Istvan~Gyongy$^2$, and, Jonathan Leach$^{1*}$.\\
$^1$School of Engineering and Physical Sciences, Heriot-Watt University, Edinburgh, EH14 4AS, UK\\
$^2$School of Engineering, The University of Edinburgh, Edinburgh, EH9 3FF, UK \\
$^*$\email{j.leach@hw.ac.uk}}

\begin{document}

\maketitle

\begin{abstract}
The growing ubiquity of drones has raised concerns over the ability of traditional air-space monitoring technologies to accurately characterise such vehicles. Here, we present a CNN using a decision tree and ensemble structure to fully characterise drones in flight.  Our system determines the drone type, orientation (in terms of pitch, roll, and yaw), and performs segmentation to classify different body parts (engines, body, and camera). We also provide a computer model for the rapid generation of large quantities of accurately labelled photo-realistic training data and demonstrate that this data is of sufficient fidelity to allow the system to accurately characterise real drones in flight. Our network will provide a valuable tool in the image processing chain where it may build upon existing drone detection technologies to provide complete drone characterisation over wide areas.
\end{abstract}

\section{Introduction}
\label{sec:introduction}
The proliferation of semi-autonomous aerial vehicles, i.e. drones, into the consumer and industrial spaces, combined with the growing number of drone related incidents (infractions into commercial airspace,~\cite{gettinger2015drone,mehta2020before} or the use of drones by militant groups,~\cite{rossiter2018drone,hammes2016democratization}) has raised concerns over the ability of existing aerial detection systems to accurately characterise such vehicles~\cite{schneider2019regulators,nguyen2019towards,o2019no}.
\begin{figure}[b!]
	\centering
        \includegraphics[width=0.85\columnwidth]{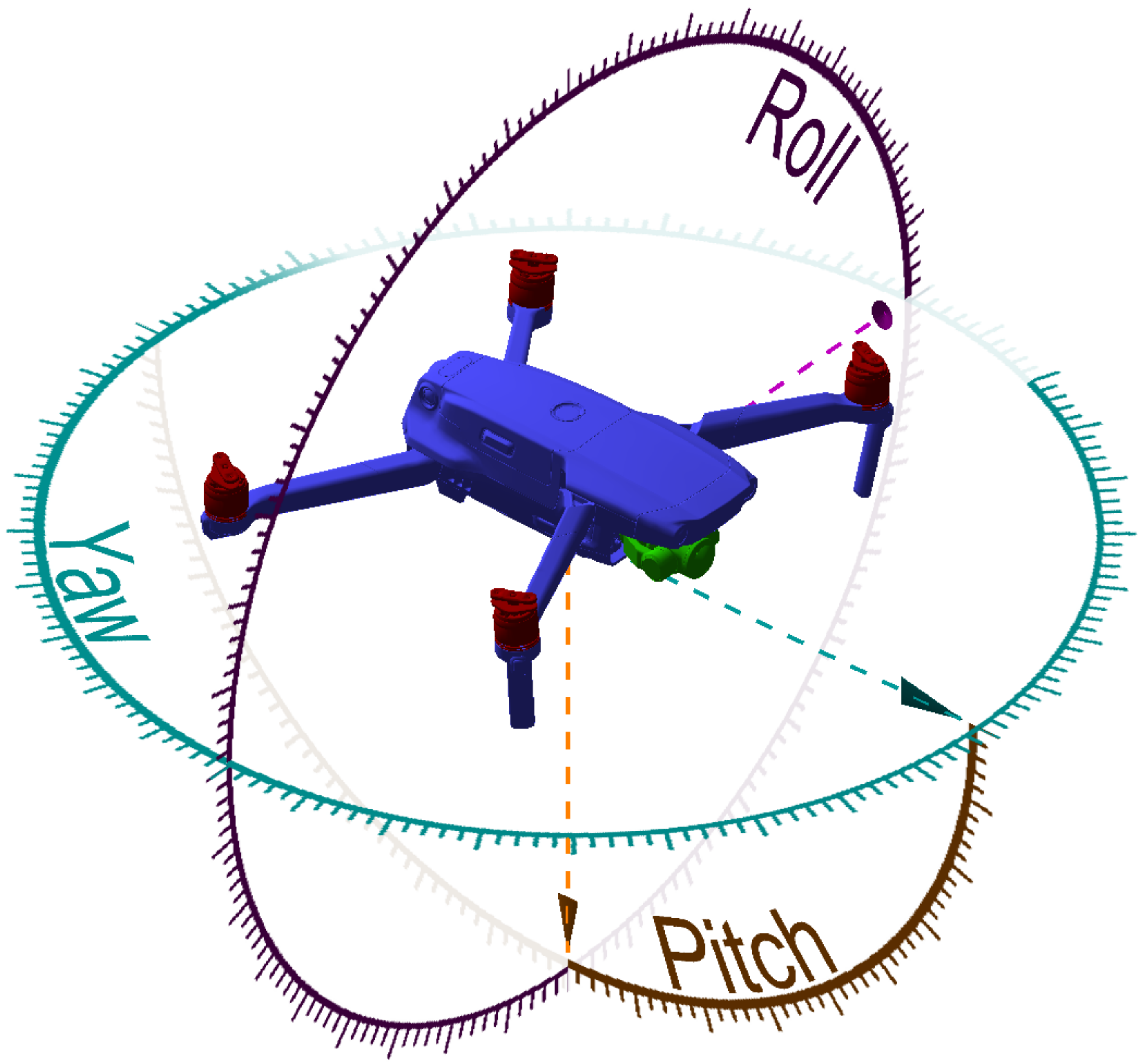}
        \caption{A conceptual representation of drone pose. A drone (here represented by a DJI Mavic 2) is identified and divided into its components,  for instance, body (blue), engines (red), and, camera (green). Further, the orientation of the drone in 3D space as represented by the roll (magenta), pitch (orange) and yaw (cyan) gimbals is identified.}
	    \label{fig: Concept}
\end{figure}
Specifically, many existing air-space monitoring technologies are optimized to detect the presence of a vehicle, identify its type, and, track its position over time but they lack the resolution to determine target specific features. This, in conjunction with drones ability to decouple their motion in space from their assigned task e.g. simultaneously translate and rotate to keep a subject in frame whilst filming, means that presence, type and position are often insufficient to accurately identify the intent of a vehicle.\\
\\
To accurately assess the intent of a drone it is necessary to fully characterize its `pose' i.e., not only identify its type but also segment it into functional components and identify the orientation of these components in 3D space. Fig.~\ref{fig: Concept} conceptually illustrates this process showing a DJI Mavic 2 drone segmented into colour-coded components and placed within a 3D Gimbal corresponding to its orientation.\\
\\
To address the problem of drone characterization a wide variety of machine learning assisted drone detection systems have been developed.  For example,  radio based methods, which eavesdrop on the communications between drones and pilots and apply the statistical analyses of control signals~\cite{RN229,RN230,RN58,RN55}, Convolutional Neural Networks (CNNs) analysing the spectragram~\cite{RN246,RN348,RN249,RN168}, K-Nearest Neighbours (KNNs)~\cite{RN171} clustering of signals, cyclostationary feature extractors~\cite{RN93}, decision trees~\cite{RN203} and random forest techniques~\cite{RN327}, bit-analysis~\cite{RN113}, and, residual~\cite{RN227}, recurrent~\cite{RN259} and hierarchical networks~\cite{RN45}. Additionally, acoustic based methods analysing the noise of a drones motors and propellers have also been developed using Mel Frequency Cepstral Coefficients (MFCC)~\cite{RN260,RN224,RN279,RN223,RN48,RN77} or by converting the signal to a spectragram~\cite{RN233,RN433,RN223}. Once obtained, the MFCC or spectragram feature set can be used to train Long-Short Term Memory (LSTM) models~\cite{RN260}, or Convolution type models such as CNNs~\cite{RN164,RN262,RN261,RN200,RN433,RN239}, Recurrent Neural Networks (RNNs)~\cite{RN164,RN261,RN262} which incorporate temporal dependence and Convolutional-RNNs (CRNNs)~\cite{RN261,RN262}. The feature set can also be used to train vector type models including, Support Vector Machines (SVMs)~\cite{RN233,RN223,RN200,RN48}, Gaussian Mixture Models~\cite{RN164} and KNNs~\cite{RN169} or, retrain existing models such as random forests~\cite{RN238}, ResNet~\cite{RN224} and LeNet~\cite{RN244}.\\
\\
Despite the relative efficacy of acoustic and radio based systems the introduction of quiet micro-drones and fully autonomous drones (which do not require radio commands) has rendered them progressively less versatile and has necessitated the development of radar and optical based sensor systems. Radar in particular has seen extensive development including pulsed systems~\cite{RN306,RN152,RN154}, Doppler systems~\cite{RN311,RN307,RN35,RN95,RN46}, and Frequency Modulated Continuous Wave (FMCW) systems~\cite{RN114,RN317,RN59} all at multiple wavelengths~\cite{RN314,RN119,RN76,RN112,RN298,RN165,RN105,RN166,RN53,RN220}. The reader is directed to Refs~\cite{RN304,RN151,RN87,RN71} for a comprehensive review. Whilst radar based systems are able to monitor a large area and are robust to atmospheric conditions, their reliance on micro-Doppler information for drone type identification and poor transverse resolution has prevented their application to problems beyond target detection and tracking. Hence, in parallel to radar systems, machine learning assisted optical drone detection systems have been developed. Such systems have been extensively used to identify the presence of drones in an image and construct bounding boxes at ranges comparable to that of radar systems~\cite{RN221,RN194}.\\    
\\
The most common approach to optical drone detection is to train existing CNN based networks such as You Only Look Once (YOLO)~\cite{redmon2016you,RN190} and ResNet~\cite{he2016deep,RN32} on colour camera images. These networks include coupling to pan-tilt and zoom camera mounts to track moving objects~\cite{RN159}, using multi-camera systems to increase the field of view~\cite{RN344,RN96}, utilising the high speed nature of YOLO to identify drones at video frame rates~\cite{RN219}, comparing the performance of YOLO v2 and YOLO v3 on drones at short range against static backgrounds~\cite{RN50}, examining the effect of incorrect images labels on YOLO~\cite{RN170} and, modified YOLO implementations~\cite{RN407}.\\
\\
More complex optical CNN architectures have also been developed where features in the image (such as moving objects) are enhanced before being sent to a second network for identification. These multi-stage networks have proven to generally be more effective at discriminating drones from drone-like objects in images such as birds~\cite{RN101,RN445,RN207,RN74}. Such networks have been developed using background subtraction with image stabilization~\cite{RN25} and CNNs~\cite{RN83,RN9}, subtracting successive frames and clustering using an SVM~\cite{RN70}, HAAR filters~\cite{RN198} for edge and feature detection, foreground background separation~\cite{RN43}, ResNet for feature extraction and SVMs for classification~\cite{RN150}, Kalman filters and ResNet~\cite{RN241}, Faster-RCNN and ResNet~\cite{RN5}, and, using trajectory mapping to suppress erroneous YOLO identifications~\cite{RN236}. Additionally, several other networks have been used for drone identification.  These include identifying regions of interest in an image~\cite{RN44} using Histogram of Gradient (HOG) descriptors with thresholding or Fourier descriptors~\cite{RN82}, simultaneous image upsampling and downsampling~\cite{RN92}, Inception Net v3~\cite{RN267}, generic Fourier descriptors~\cite{RN111,RN455}, Faster-RCNN~\cite{RN69}, and, TIB-Net with CenterNet, lightweight networks optimised for speed of processing~\cite{RN72,RN257}.\\ 
\\
Finally, a number of more niche applications have also been investigated such as, controlling the flight of a drone based on external camera observations~\cite{RN272} and, using cameras mounted on multiple drones to track and even intercept hostile drones~\cite{RN157,RN450,RN86,RN258}. For a review of the different machine learning implementations listed above the reader is directed to Refs~\cite{RN17,RN418,RN256}. Despite the numerous optical systems developed to date characterisation of drones beyond presence, location and type remains rare with demonstrations limited to determining if a drone is carrying a payload~\cite{RN31} or the identification of key points on a single drone at short range~\cite{RN91}.\\
\\
A promising avenue for the more complete characterization of drones is given by sensor fusion in which multiple sensors are combined. For example, using a large field-of-view low resolution sensor to direct a small field-of-view high resolution sensor with such systems seeing improvements in performance of up to 15\%~\cite{RN286,RN344,RN68}. In the case of optical drone detection systems one such example is the development of depth sensing time-of-flight systems such as LIDARS. LIDARS active illumination allows them to operate when no passive light source is available (such as at night), detect targets which themselves emit no thermal radiation, and, operate to a limited degree through obscurance. Scanning LIDARS have been shown to be effective at drone detection at ranges up to 2 km when coupled with a Variable Radially Bounded Nearest Neighbour (V-RBNN) network to analyse the point cloud~\cite{RN336,RN135}. Further, flash LIDAR systems such as those employing Single Photon Avalanche Detector (SPAD) array cameras allow for the simultaneous capture of a `traditional' high transverse resolution intensity image as well as a lower transverse resolution depth image (where `depth' refers to the distance between the camera and the object for each pixel). Such systems have been shown to be effective at identifying the pose of objects at short range~\cite{gyongy2020high,mora2021high} but have yet to be applied to the problem of drone characterization.\\   
\\
Here, we present a CNN which provides the complete characterization of drones. The network takes as input an intensity image as well as depth data and outputs: the identity of the drone i.e., the type of drone in the data; the segmentation of the drone in which each pixel in the intensity image is classified according to the drone component it represents; and, the orientation, the angle of the drone about its three principle axes of rotation, yaw, pitch, and, roll. We examine the performance of the network in multiple scenarios including, different drones, different ranges of motion and different data inputs. Additionally, we outline a system for producing large quantities of accurately labelled simulated data on which we train our network. To verify both our network structure and our simulated training data we demonstrate the ability of our network to accurately characterize an image of a real DJI Mavic 2 Zoom drone in flight as captured by a Quantic4x4 SPAD camera~\cite{hutchings2019reconfigurable}. The SPAD camera represents a state-of-the-art sensor fusion system combining a functional transverse resolution of 80$\times$240 pixels for intensity and 20$\times$60 pixels for depth. Further, each depth pixel outputs a depth histogram with 500 picosecond temporal resolution. Finally, the architecture of the chip has the potential for the alternating acquisition of visible spectrum intensity images and depth histograms at rates in excess of 1000 frames per second \cite{gyongy2020high}.        

\section{Network Architecture}
\label{sec:network_architecture}
We present a network architecture built on a decision tree coupled with an ensemble network. The decision tree identifies the type of drone after which a set of drone-specific pretrained networks are applied in parallel to perform the orientation and segmentation operations. Specifically, the orientation is determined by three identical networks each trained to identify a single axis (roll, pitch or yaw) while the segmentation is performed by an additional U-Net~\cite{ronneberger2015u} type network. This structure allows multiple drone parameters to be identified simultaneously through network parrallalization whilst allowing each network to be optimized on a specific parameter yielding superior overall performance.\\
\begin{figure}[t!]
	\centering
	    \setlength\tabcolsep{1pt}
	    \footnotesize
        \begin{tabular}{  c c c  }
             \multicolumn{1}{c}{\bfseries   }&\multicolumn{1}{c}{\bfseries Mavic 2}& \multicolumn{1}{c}{\bfseries Inspire 2}\\

            \cmidrule(lr){2-2}
            \cmidrule(lr){3-3}
            \begin{tabular}[t]{@{}c@{}}a)\end{tabular}
              &\multirow{2}{*}{\includegraphics[width=0.46\columnwidth]{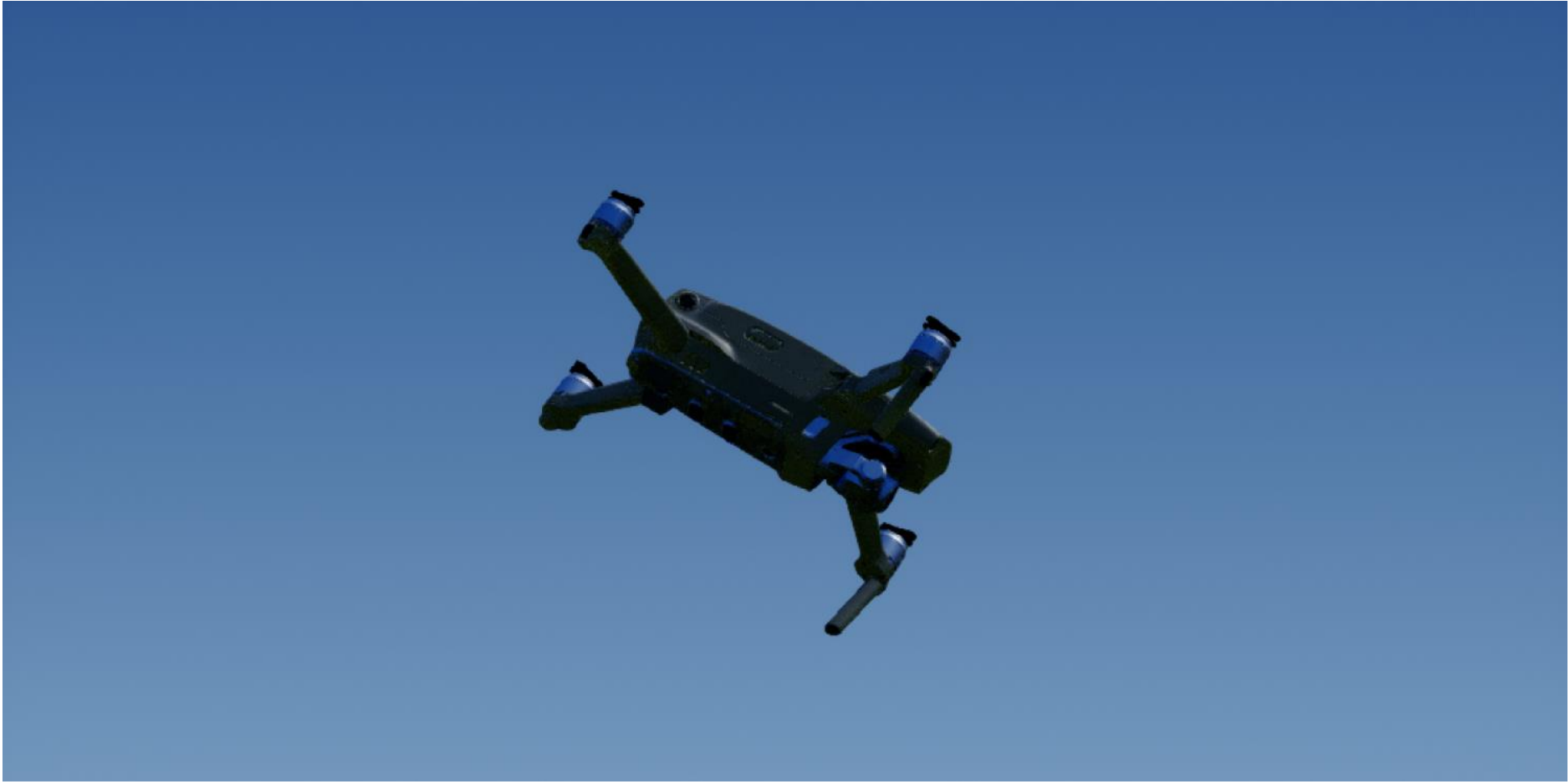}} & \multirow{2}{*}{\includegraphics[width=0.46\columnwidth]{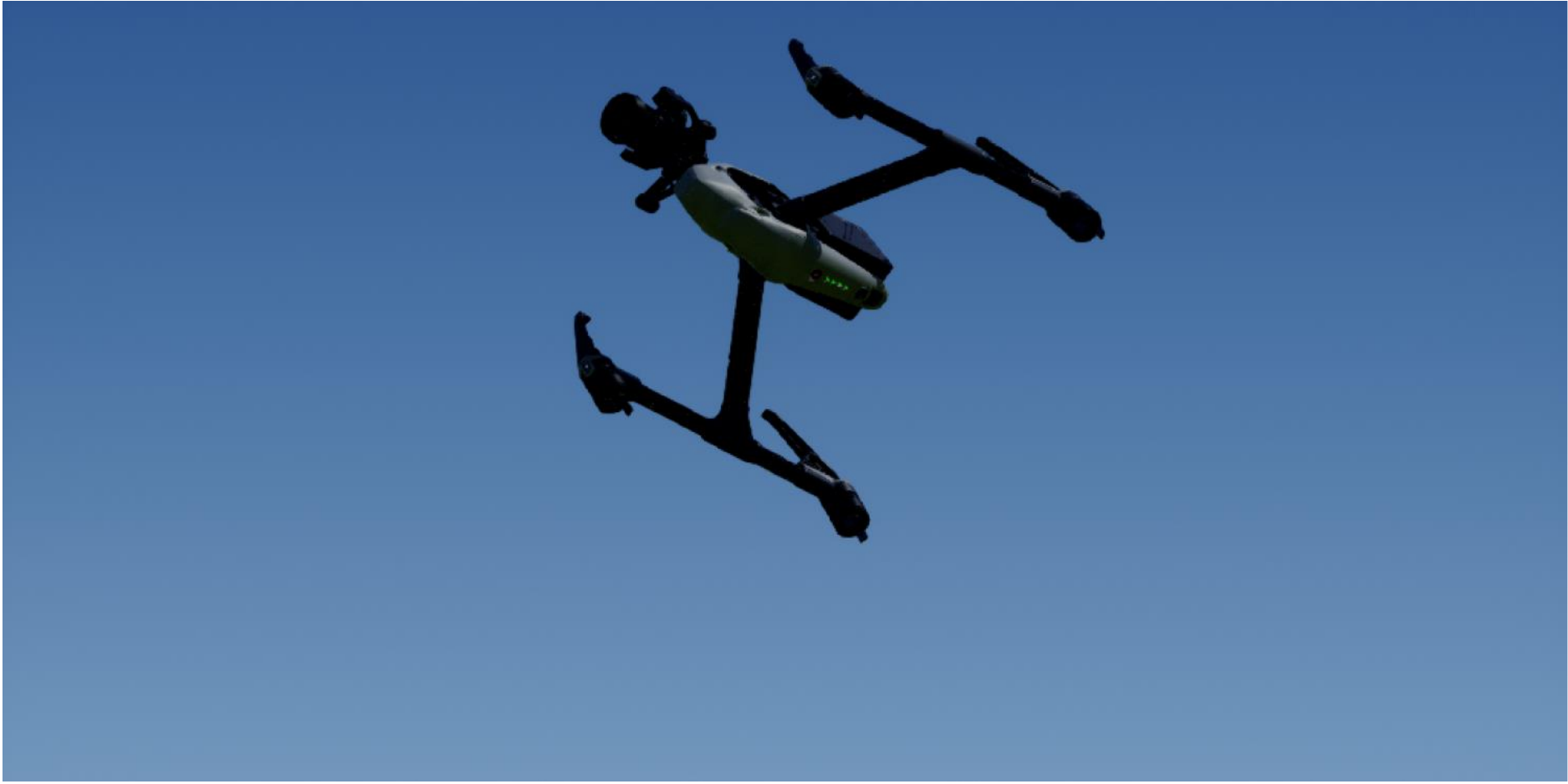}}\vspace{1.4cm}\\
             &&\\
     
            \cmidrule(lr){2-2}
            \cmidrule(lr){3-3}
            \begin{tabular}[t]{@{}c@{}}b)\end{tabular}
            
             &\multirow{2}{*}{\includegraphics[width=0.46\columnwidth]{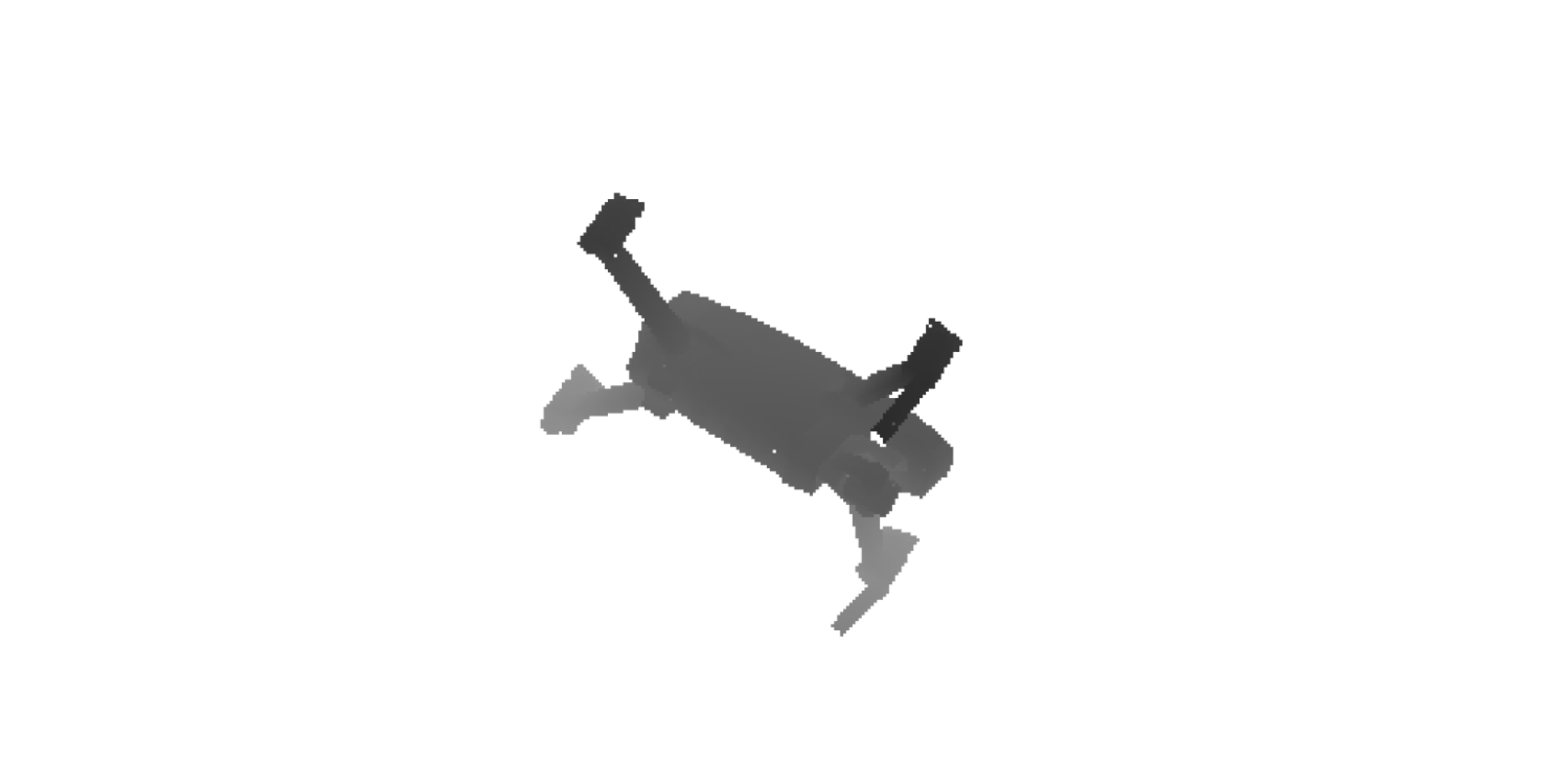}} &
             \multirow{2}{*}{\includegraphics[width=0.46\columnwidth]{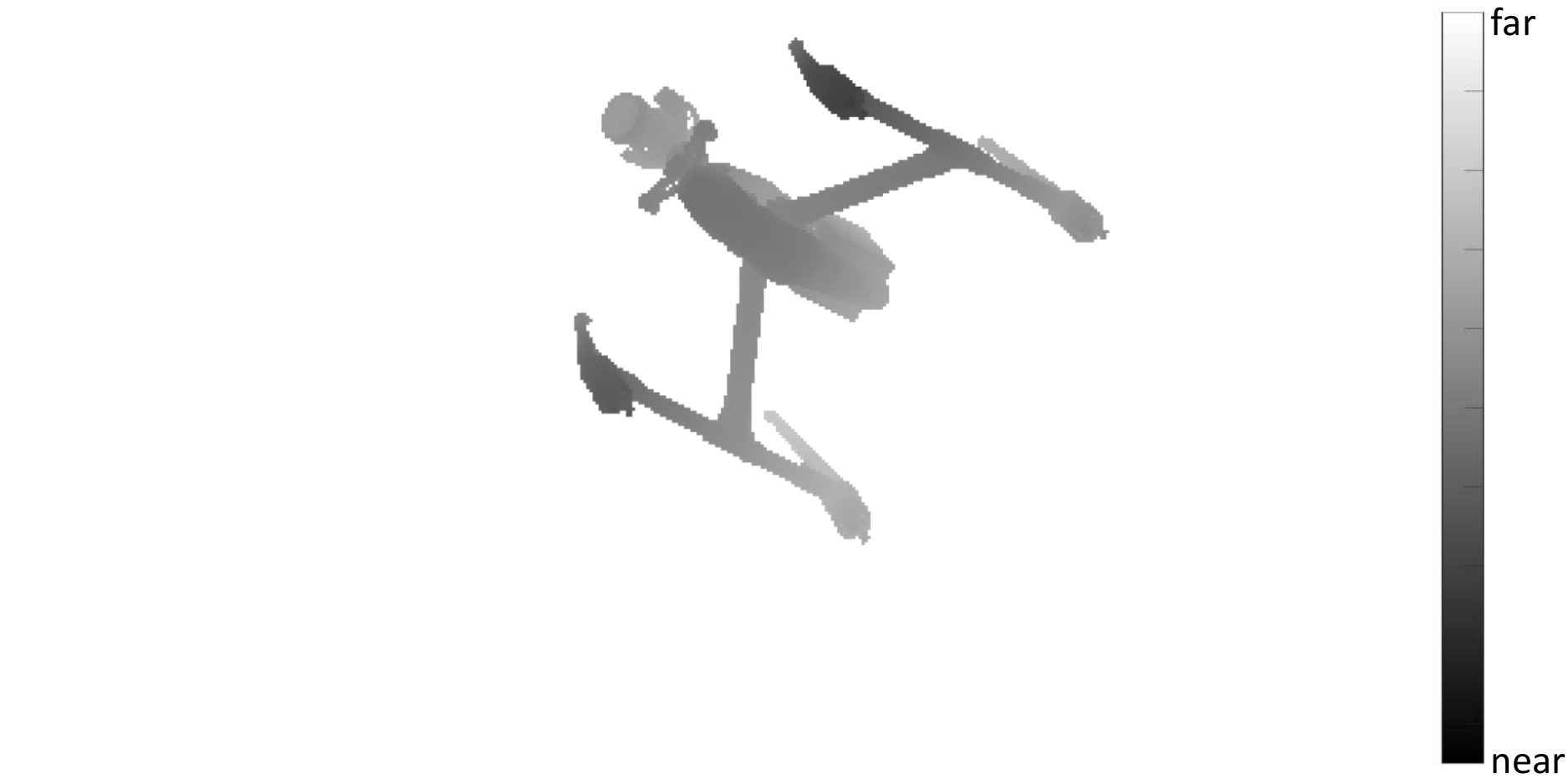}}\vspace{1.4cm}\\
             &&\\
   
            \cmidrule(lr){2-2}
            \cmidrule(lr){3-3}
            \begin{tabular}[t]{@{}c@{}}c)\end{tabular}
            
             &\multirow{2}{*}{\includegraphics[width=0.46\columnwidth]{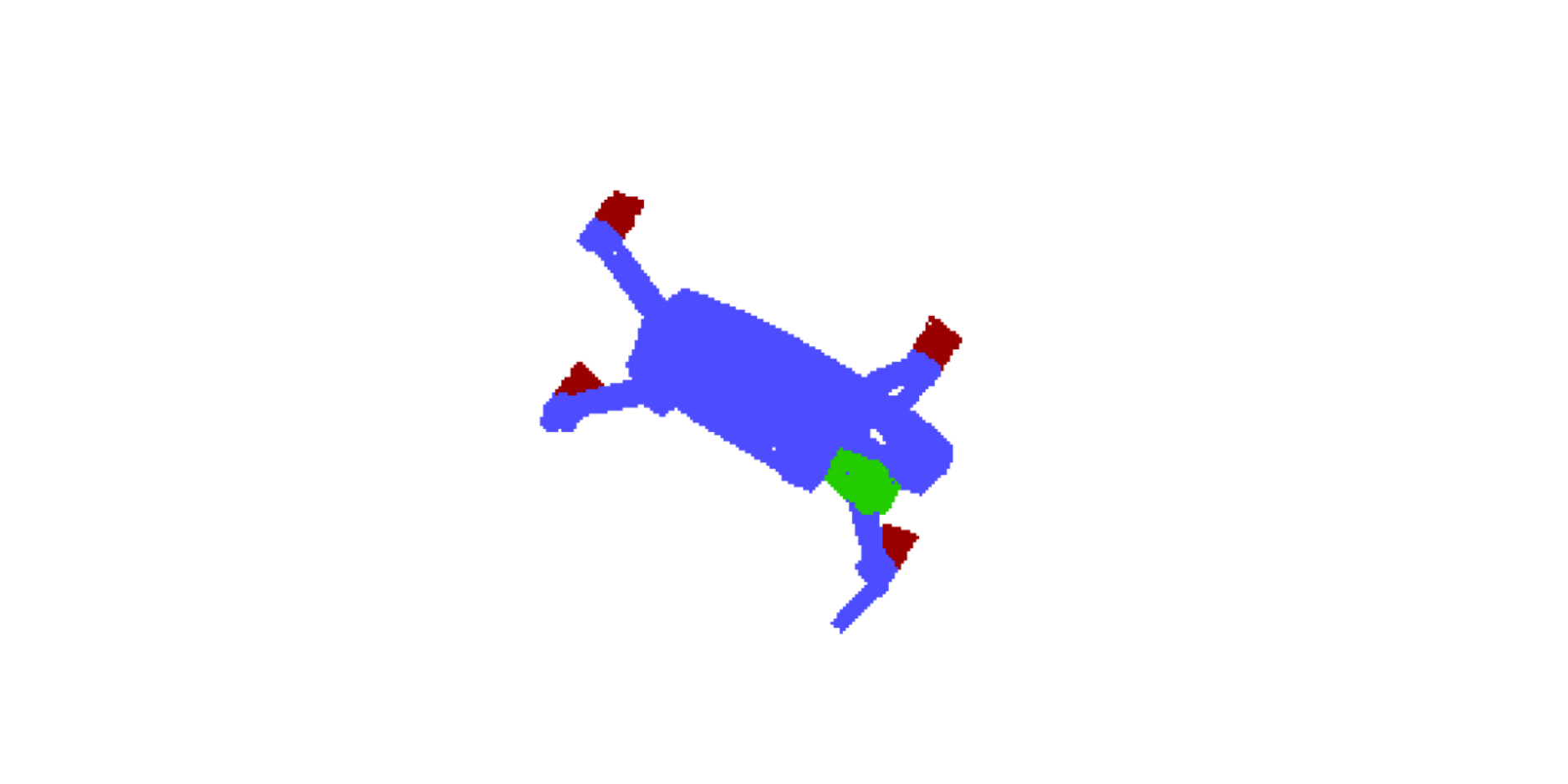}} & 
             \multirow{2}{*}{\includegraphics[width=0.46\columnwidth]{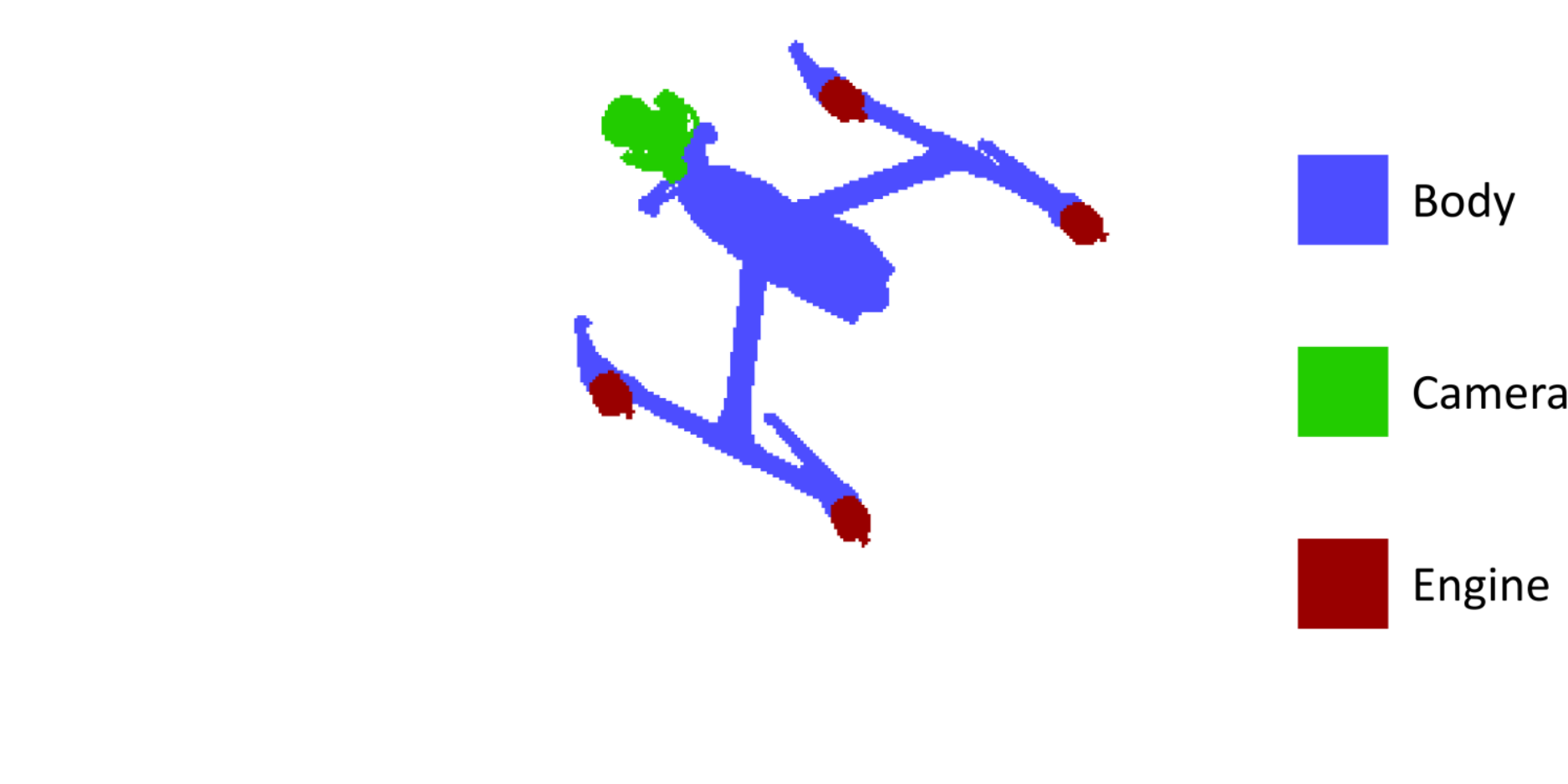}}\vspace{1.1cm}\\
             &&\\
        \end{tabular}
        \captionof{figure}{Examples of the Unreal engines ability to produce realistic, accurately labelled, intensity, depth and segmentation data. a) Intensity images generated by the Unreal engine of a DJI Mavic 2 and (an upside down) DJI Inspire 2 drone in flight. b) Unreal engine depth images corresponding to the drones in the top panels. c) Segmentation labels from the Unreal engine for the drones in the top panels showing the body (in blue), the engines (in red), and, the cameras (in green).}
	    \label{fig: Unreal_high_res}
    \end{figure}\\
\begin{figure}[b!]
	\centering
	    \setlength\tabcolsep{1pt}
	    \footnotesize
        \begin{tabular}{  c c c  }
             \multicolumn{1}{c}{\bfseries   }&\multicolumn{1}{c}{\bfseries Intensity}& \multicolumn{1}{c}{\bfseries Depth}\\

            \cmidrule(lr){2-2}
            \cmidrule(lr){3-3}
            \begin{tabular}[t]{@{}c@{}}a)\end{tabular}
              &\multirow{2}{*}{\includegraphics[width=0.46\columnwidth]{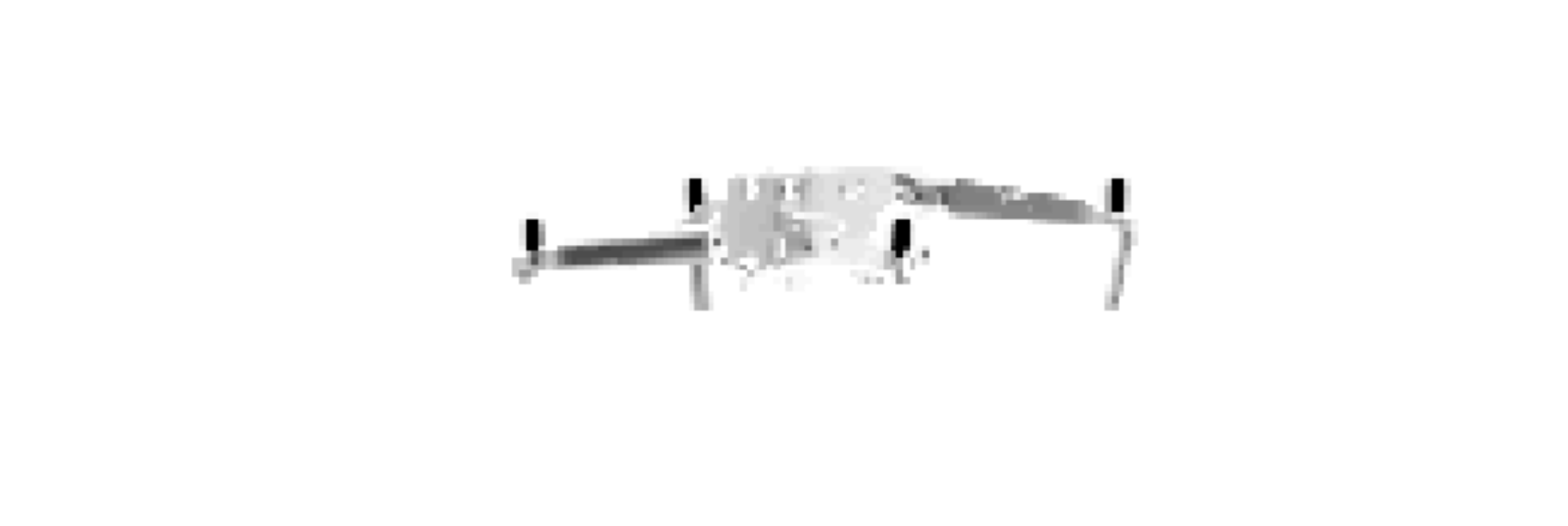}} & \multirow{2}{*}{\includegraphics[width=0.46\columnwidth]{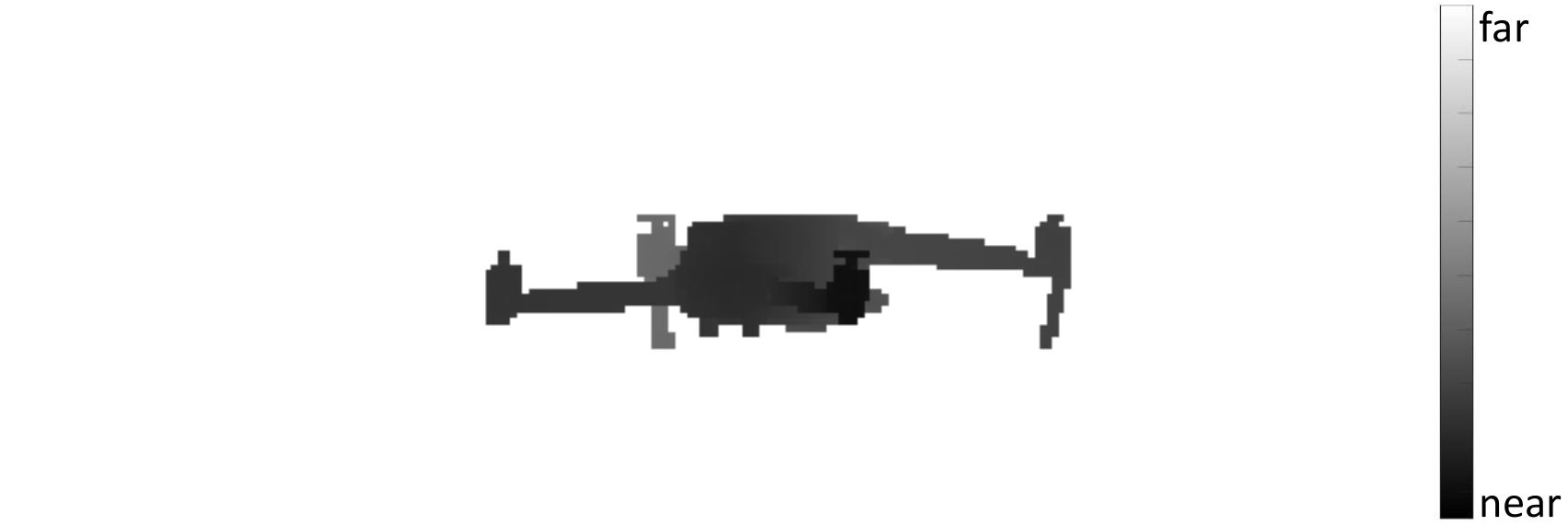}}\vspace{0.7cm}\\
             &&\\
     
            \cmidrule(lr){2-2}
            \cmidrule(lr){3-3}
            \begin{tabular}[t]{@{}c@{}}b)\end{tabular}
            
             &\multirow{2}{*}{\includegraphics[width=0.46\columnwidth]{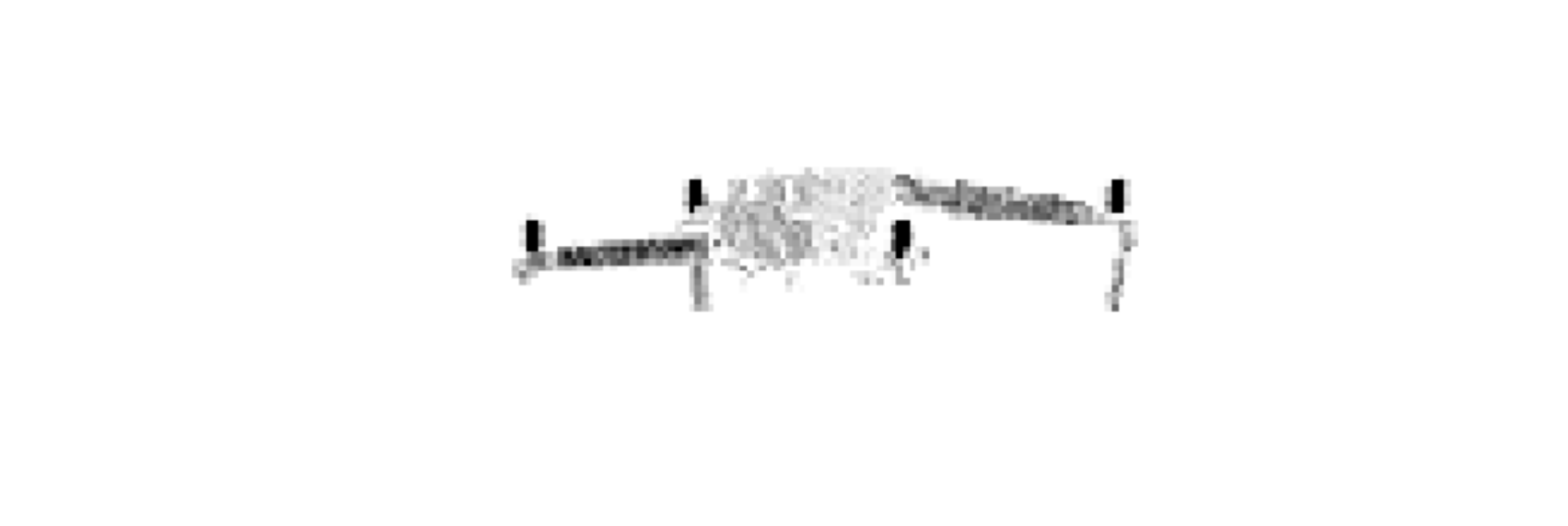}} &
             \multirow{2}{*}{\includegraphics[width=0.46\columnwidth]{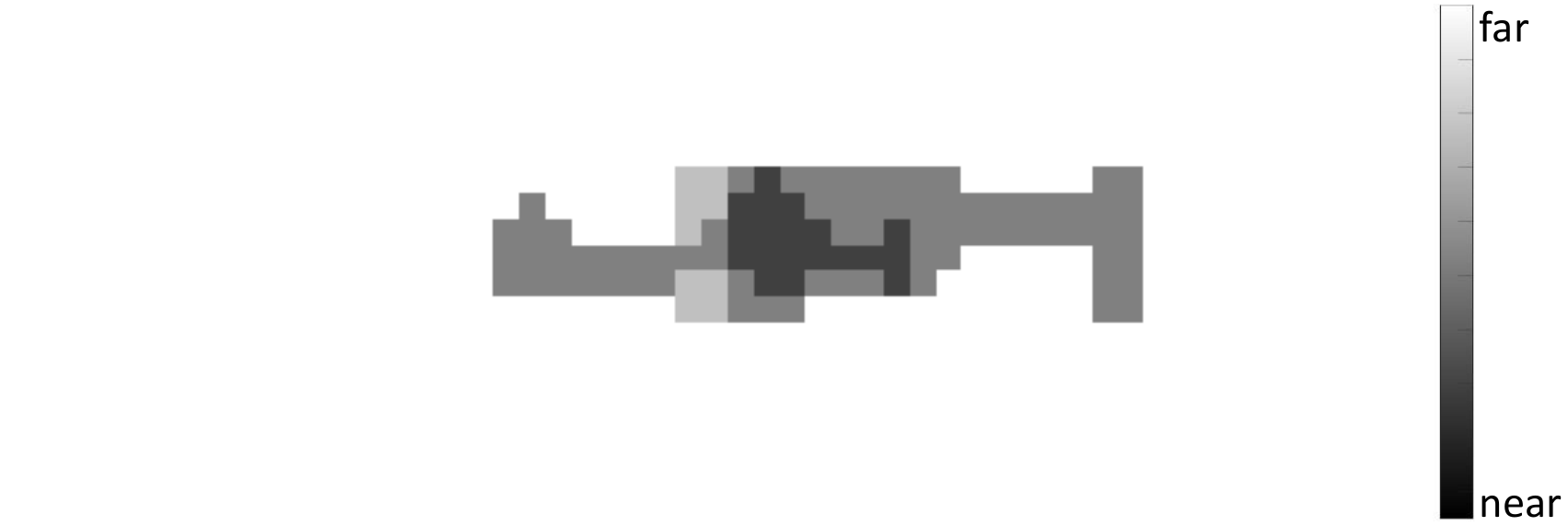}}\vspace{0.7cm}\\
             &&\\
   
            \cmidrule(lr){2-2}
            \cmidrule(lr){3-3}
            \begin{tabular}[t]{@{}c@{}}c)\end{tabular}
            
             &\multirow{2}{*}{\includegraphics[width=0.46\columnwidth]{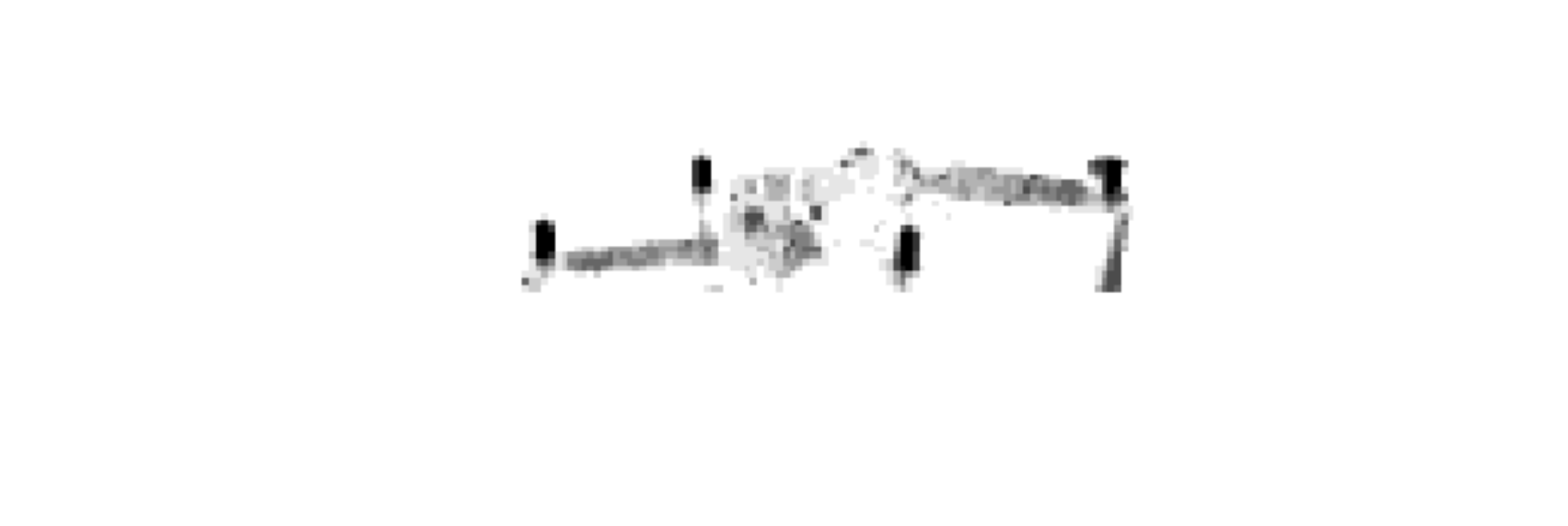}} & 
             \multirow{2}{*}{\includegraphics[width=0.46\columnwidth]{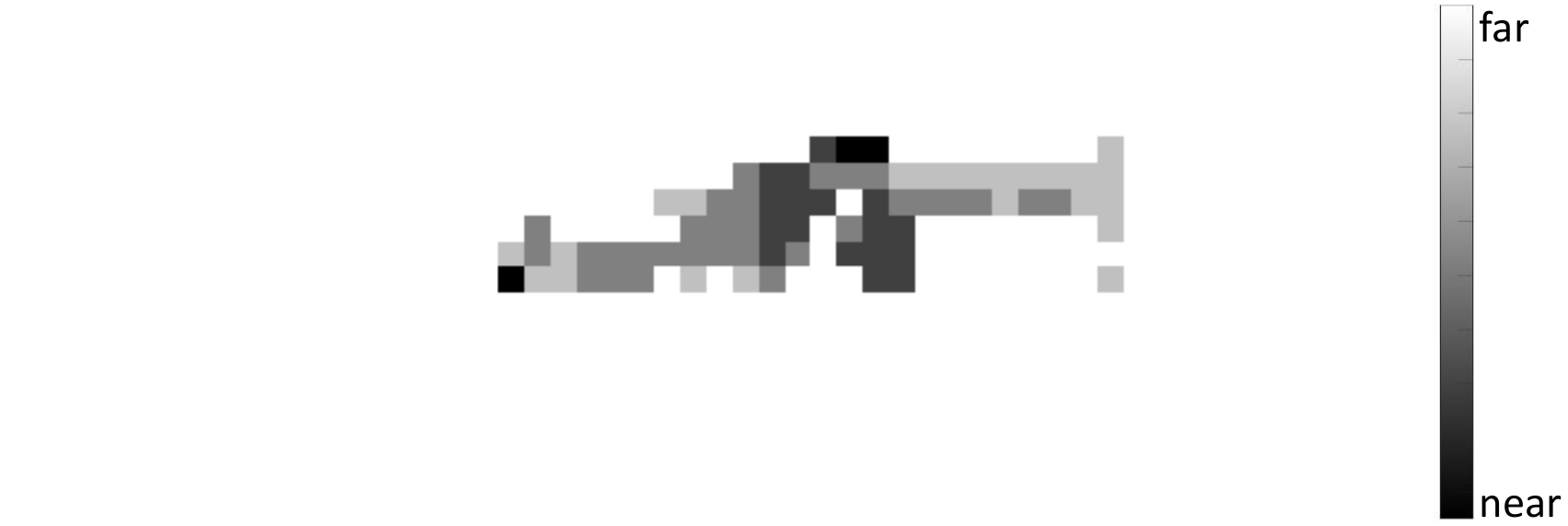}}\vspace{0.7cm}\\
             &&\\
        \end{tabular}
        \captionof{figure}{Processed images from Unreal Engine compared to Quantic4x4 SPAD camera images. a) The intensity and depth images produced by the Unreal environment. b) The data used to train the network. The intensity image is noised with a Poisson filter while the depth image is down-sampled and converted to a histogram of depths (visualised here as a depth image). c) Images captured by a Quantic4x4 SPAD camera of a real drone in flight. Note that the intensity images have been enhanced in contrast for better visualization.}
	    \label{fig: Network inputs}
    \end{figure}\\
The lack of high quality drone image training datasets remains an obstacle for machine learning assisted drone classification. To address this, several publications have examined data augmentation~\cite{RN11} techniques such as, super-imposing drone images onto unrelated backgrounds~\cite{RN347}, super-resolution upscaling~\cite{RN49} and, generating new images from Generational-Adverserial Networks (GANs)~\cite{RN139}. Here, we leverage the capability of the Unreal Engine video game development environment to rapidly produce a large set of photo-realistic, accurately labelled training data as illustrated by Fig.~\ref{fig: Unreal_high_res}. This approach allows us to explore the parameter space of drone types, orientation limits (e.g. the upside down Inspire 2 in Fig.~\ref{fig: Unreal_high_res}), lighting conditions and image qualities to an extent which would be impractical experimentally. The Unreal code is publicly available and can be found at https://github.com/HWQuantum/.
\begin{figure}[t!]
	\centering
        \includegraphics[width=\columnwidth]{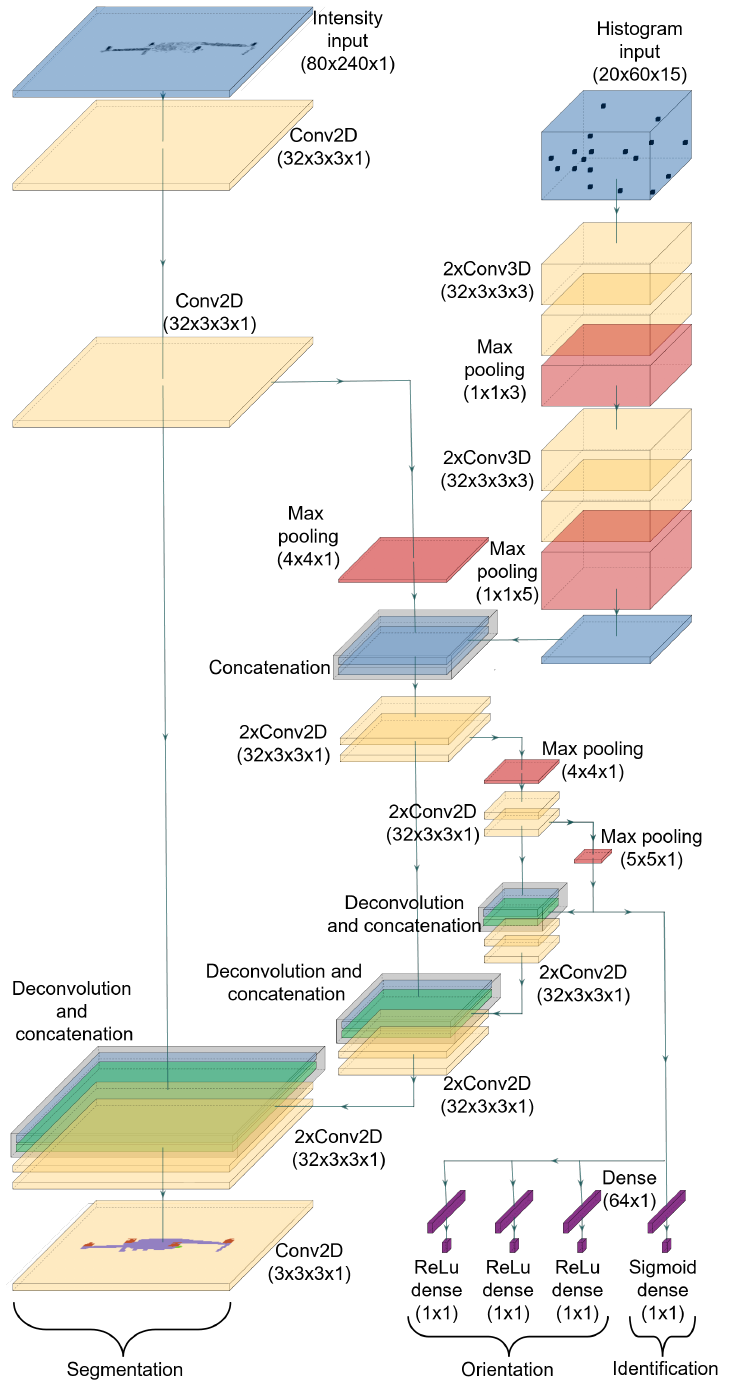}
        \caption{A summary of the ensemble network structure, here the common components of the networks have been drawn together while in practice each network in the ensemble is distinct. The networks take in a high transverse resolution intensity image and a low transverse resolution histogram of depth. Using convolution, pooling, and, concatenation the inputs are reduced to a dense latent space. The identification network connects this latent space to a dense layer and then to a single Sigmoid activated neuron for drone type classification. By contrast, the three orientation networks use an identical structure but employ ReLu activation in the final neuron to output a continuous value corresponding to the angle in a given axis. Segmentation is performed by up-sampling the latent space to a final convolution with filters corresponding to the components being identified.}
	    \label{fig: CNN}
\end{figure}
Fig.~\ref{fig: Network inputs} shows examples of the images processed by the network. The simulated images produced by the Unreal environment (Fig.~\ref{fig: Network inputs}a)) are processed before being passed to the network. The simulated intensity image is noised with a Poisson filter (Fig.~\ref{fig: Network inputs}b)) and resized to 80$\times$240 pixels, while the depth is downsampled and converted to a histogram with a dimensionality of 20$\times$60$\times$15. Fig.~\ref{fig: Network inputs}c) shows the images produced by a Quantic4x4 SPAD array camera of a real drone in flight which the simulated data is designed to mimic. We stress that the image sizes used in the simulated data were selected only to match the physical parameters of the QuantIC4x4 SPAD sensor, and the network can be reshaped to any dataset with both intensity and depth information.\\
\\
Fig.~\ref{fig: CNN} shows a summary of the identification, orientation and segmentaion networks. At the core of these networks is the Drone Feature Encoder (DFE) which reduces the input data to a latent feature space. The DFE takes as input both a histogram of depths (of size $r_h$ rows, $c_h$ columns, and $p_h$ pages) and an intensity image (of size $(r_i,c_i)$).  The histogram is passed twice through two 3D convolutional layers (each with 32 filters) and axial max-poolings to extract its depth features and reduce it to a dimensionality of $(r_h,c_h,1)$. The intensity image is passed through two 2D convolution layers (each with 32 filters) and a max-pooling such that it is reduced to a dimensionality of $(r_h,c_h,1)$. The intensity and depth tensors are then concatenated and passed twice through a set of two 2D convolutions (each with 32 filters) and max-poolings ultimately reducing the network inputs to a latent space of 1$\times$3$\times$32 filters. The DFE is identical in all the networks with each network distinguished by how it handles the data in this latent space.\\ 
\\
In the case of the identification network which defines the decision tree, the latent space is flattened to a 96 element vector and connected to a dense layer with 64 neurons. These neurons are in turn connected to the single output node with a Sigmoid activation. This network is trained using cross-entropy as a loss function such that it outputs an integer corresponding to the type of drone in the image. The orientation networks are identical in structure to the identification network, but the final neuron uses a ReLu activation. ReLu activation allows the neuron to output a continuous value corresponding to the angle in a given axis. Further, the orientation networks are trained using the loss function given in Eqn.~\ref{eqn:Loss_func} which allows them to correctly account for the cyclic nature of angle prediction.\\
\begin{equation}
    \begin{aligned}
        \text{Loss} = \text{min}[\text{abs}(l-p),\text{abs}(l-p-360^\circ)]^2,
        \label{eqn:Loss_func}
    \end{aligned}
\end{equation}\\
where $l$ is the label, $p$ is the networks prediction, and, $\text{abs}$ is the absolute value function. The segmentation network attaches a U-Net to the DFE. This U-Net up-samples the latent space to a set of segmentation predictions of size $(r_i,c_i,n)$ where $n$ corresponds to the number of components being identified. Each layer of the U-Net mirrors the DFE, undoing the max-pooling and using skip connections to concatenate the tensors. These concatenated tensors are then passed through two 2D convolutions each with 32 filters. The network was trained using categorical cross-entropy. The final output is a single convolutional layer with (in this case) three filters corresponding to the three components being identified; the body of the drone, the engines of the drone, and, the camera on the drone.
\section{Results}
\label{sec:results}
\subsection{Results on simulated data}
\label{subsec:results on simulated data}
Two drones were used for testing, a DJI Mavic 2 Zoom and a DJI Inspire 2. High fidelity models of these drones were placed in the Unreal environment and a total of 72 000 simulated SPAD images generated. From these images 10\% were reserved for network testing. The networks were trained until the loss converged and the networks with the best performance on the testing data saved. These models were then validated on a set of 3600 unseen validation images. A summary of the results for the identification, segmentation and orientation networks is presented in Table.~\ref{tab: Results_Summary}.
\begin{table}[t!]
    \caption{Summary of network prediction accuracies for both drones in the full angle and reduced angle regimes.}
	\centering
	\footnotesize
        \begin{tabular}{c c c}
            \toprule
             &  \multicolumn{2}{c}{\bfseries  Mavic 2} \\ 
            \cmidrule[\heavyrulewidth](lr){2-3}
           \textbf{ Metric} & \textbf{ Full angle} & \textbf{ Reduced angle} \\
            
            \cmidrule[\heavyrulewidth](lr){1-1}
            \cmidrule[\heavyrulewidth](lr){2-2}
            \cmidrule[\heavyrulewidth](lr){3-3}

            Orientation & (accuracy $\pm$ std)(\%)  & (accuracy $\pm$ std)(\%) \\
	        \cmidrule(lr){1-1}
            
	         Roll &   88.3 $\pm$ 13.0& 99.6 $\pm$ 0.4\\
	        \cmidrule(lr){2-2}
	        \cmidrule(lr){3-3}
	 
	        Pitch &   99.2 $\pm$ 0.8& 99.7 $\pm$ 0.3\\
	        
	        \cmidrule(lr){2-2}
	        \cmidrule(lr){3-3}
	 
	        Yaw &   92.3 $\pm$ 9.7& 96.3 $\pm$ 5.0\\
	        \cmidrule(lr){1-1}
	        \cmidrule(lr){2-2}
	        \cmidrule(lr){3-3}
	 
            Segmentation &   &   \\
	        \cmidrule(lr){1-1}
            Body &   97 $\pm$ 1& 96 $\pm$ 1\\
	        \cmidrule(lr){2-2}
	        \cmidrule(lr){3-3}

	        Engine &   86 $\pm$ 9& 86 $\pm$ 5\\
	        \cmidrule(lr){2-2}
	        \cmidrule(lr){3-3}

	        Camera &   85 $\pm$ 13& 85 $\pm$ 16\\
	        \cmidrule(lr){1-1}
	        \cmidrule(lr){2-2}
	        \cmidrule(lr){3-3}

	        Identification &  100 $\pm$ 0 & 100 $\pm$ 0\\
	        \cmidrule[\heavyrulewidth](lr){1-3}
	        & \multicolumn{2}{c}{\bfseries  Inspire 2}  \\ 
            \cmidrule[\heavyrulewidth](lr){2-3}
           \textbf{ Metric} & \textbf{ Full angle} & \textbf{ Reduced angle} \\
            
            \cmidrule[\heavyrulewidth](lr){1-1}
            \cmidrule[\heavyrulewidth](lr){2-2}
            \cmidrule[\heavyrulewidth](lr){3-3}
            Orientation & (accuracy $\pm$ std)(\%)  & (accuracy $\pm$ std)(\%) \\
	        \cmidrule(lr){1-1}
            
	        Roll &  95.3 $\pm$ 5.6 & 98.8 $\pm$ 0.9\\
	        \cmidrule(lr){2-2}
	        \cmidrule(lr){3-3}

	        Pitch &  99.6 $\pm$ 0.4 & 97.5 $\pm$ 1.4\\
	        \cmidrule(lr){2-2}
	        \cmidrule(lr){3-3}

            Yaw &  92.4 $\pm$ 8.5& 96.0 $\pm$ 4.1 \\
	        \cmidrule(lr){1-1}
	        \cmidrule(lr){2-2}
	        \cmidrule(lr){3-3}

            Segmentation &   &    \\
	        \cmidrule(lr){1-1}
            Body &  91 $\pm$ 2 & 92 $\pm$ 2\\
	        \cmidrule(lr){2-2}
	        \cmidrule(lr){3-3}

	        Engine &  82 $\pm$ 9 & 84 $\pm$ 5\\
	        \cmidrule(lr){2-2}
	        \cmidrule(lr){3-3}

	        Camera &  81 $\pm$ 10 & 88 $\pm$ 6\\
	        \cmidrule(lr){1-1}
	        \cmidrule(lr){2-2}
	        \cmidrule(lr){3-3}

	        Identification &  100 $\pm$ 0 & 100 $\pm$ 0\\
	        \bottomrule
        \end{tabular}
	    \label{tab: Results_Summary}
    \end{table}\\
\\
To ensure non-negative angular values in all drone orientations a coordinate system was established in which level flight facing away from the camera corresponded to, yaw $= 180^{\circ}$, roll $=180^{\circ}$ and pitch $= 90^{\circ}$. Within this coordinate system two angular regimes were examined, the `full angle' regime and the `reduced angle' regime. In the full angle regime the drone models had the following range of motion: yaw $\in [0^{\circ},360^{\circ}]$; roll $\in[0^{\circ} ,360^{\circ}]$; and, pitch $\in [0^{\circ},180^{\circ}]$ (with pitch limited to $[0^{\circ},180^{\circ}]$ to negate gimbal-lock). In the reduced angle regime the models were constrained to within the manufacturer's limits specifically: yaw $\in [0^{\circ},360^{\circ}]$; roll $\in[140^{\circ},220^{\circ}]$; and, pitch $\in [140^{\circ},220^{\circ}]$. By examining these regimes we evaluate the network's ability to characterise drones flying in both conventional flight modes and more exotic flight modes (such as upside down).\\

\begin{figure}[b!]
	\centering
	    \footnotesize
        \begin{tabular}{ c c c  }
             \multicolumn{1}{c}{\bfseries   Axis}&\multicolumn{1}{c}{\bfseries   Mavic 2}& \multicolumn{1}{c}{\bfseries   Inspire 2}\\
             
             \cmidrule(lr){1-1}
            \cmidrule(lr){2-2}
            \cmidrule(lr){3-3}
            \multicolumn{1}{c}{ Roll}&&\\
              &\includegraphics[width=0.4\columnwidth]{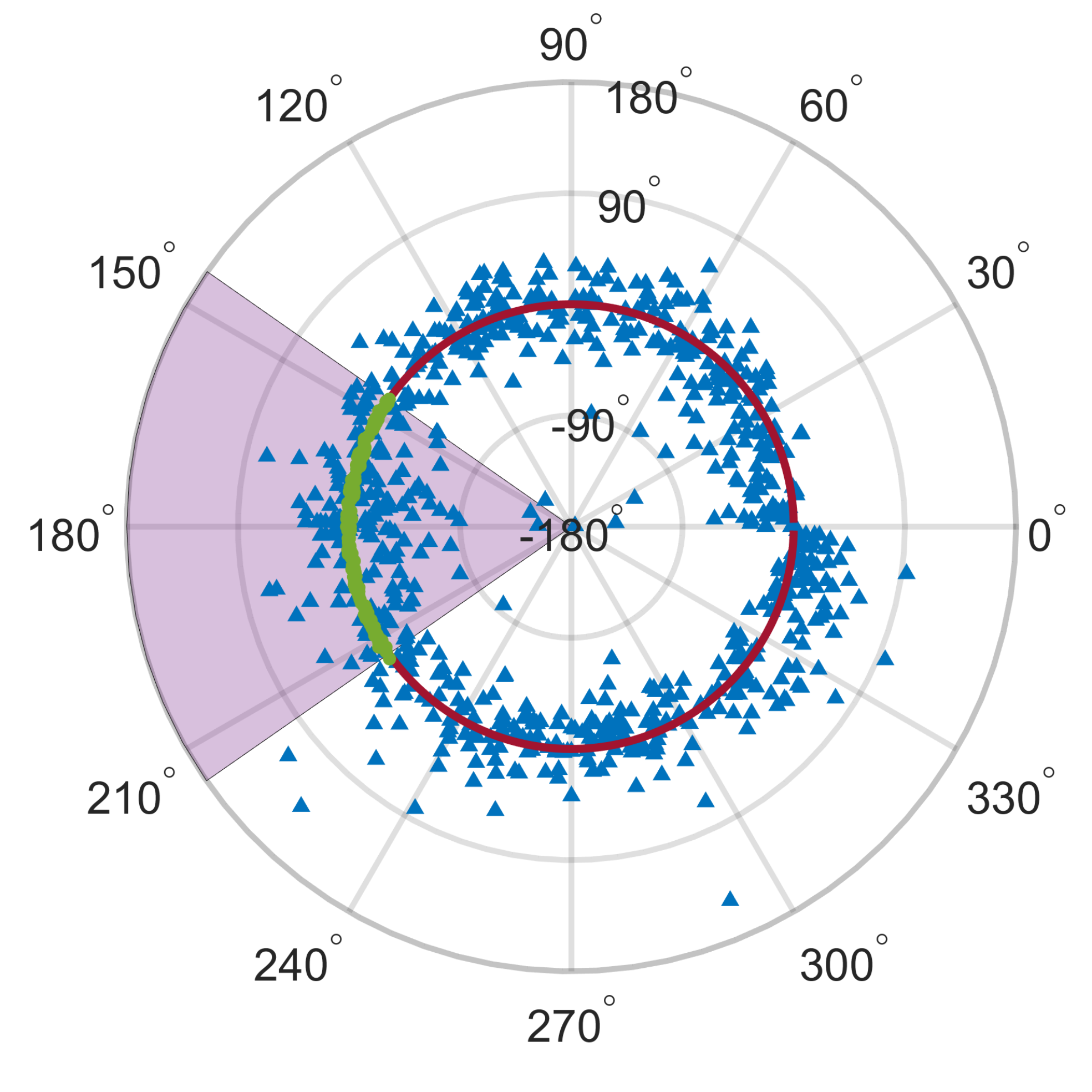} & \includegraphics[width=0.4\columnwidth]{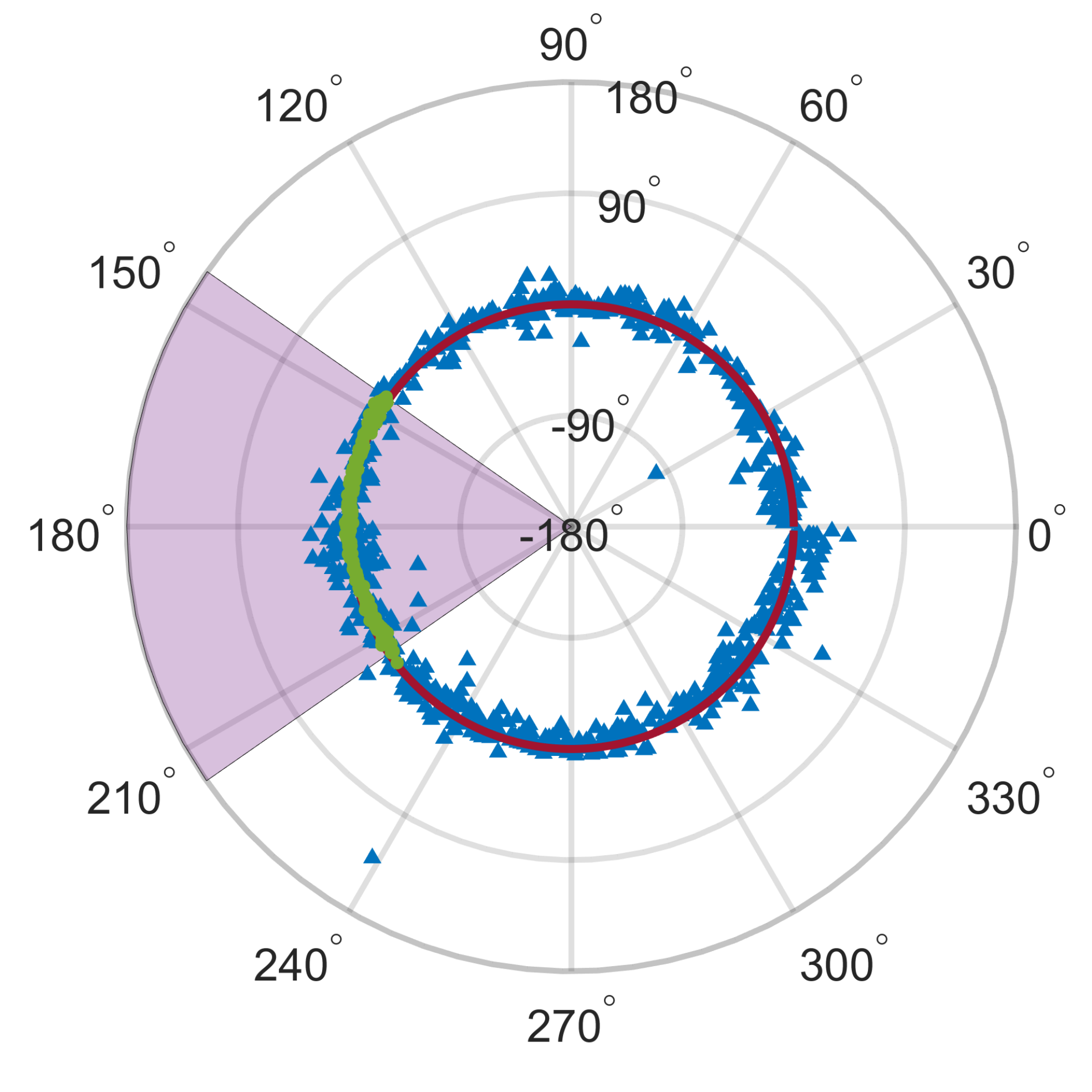}\\
             
             \cmidrule(lr){1-1}
            \cmidrule(lr){2-2}
            \cmidrule(lr){3-3}
            \multicolumn{1}{c}{ Pitch}&&\\
            
             &\includegraphics[width=0.4\columnwidth]{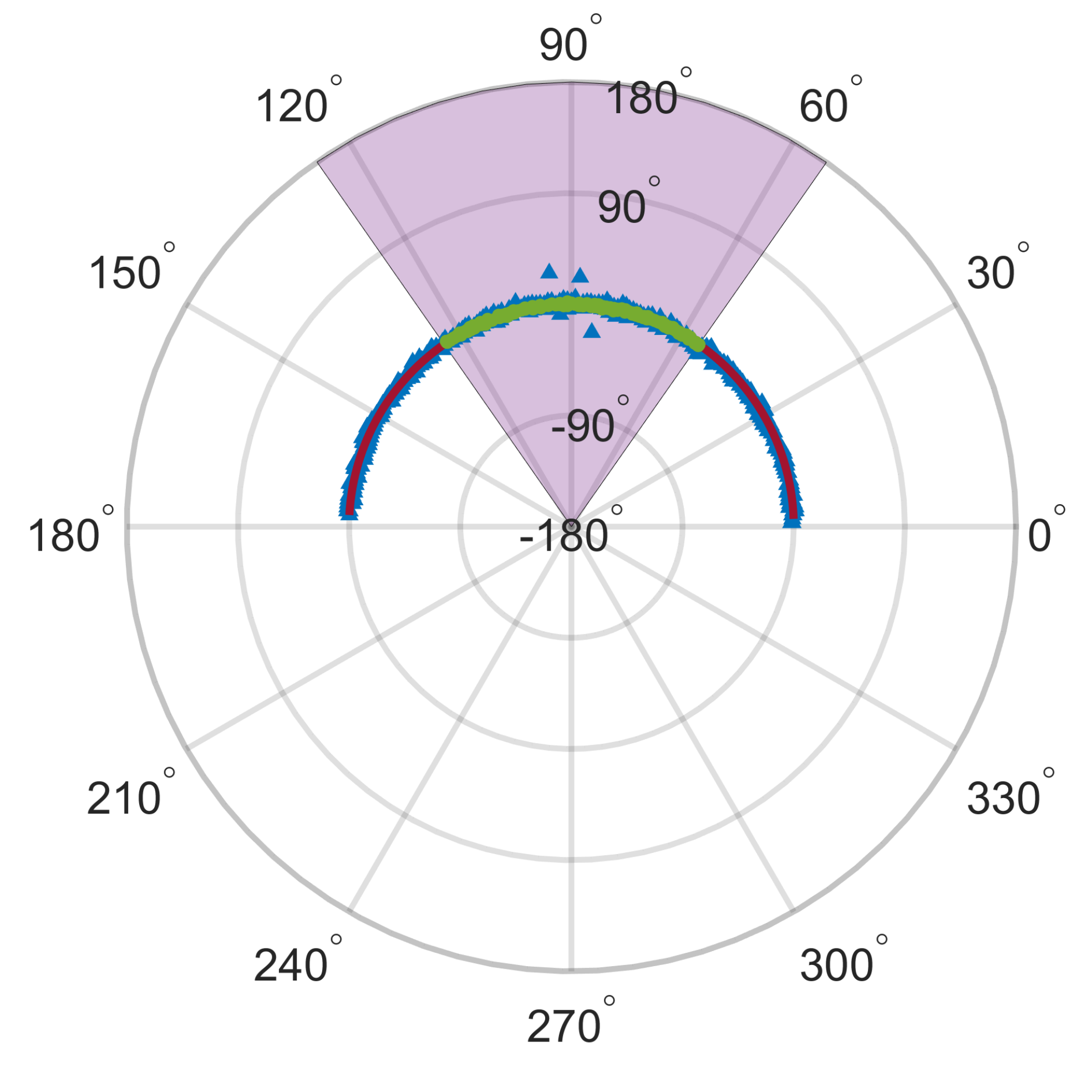} &
             \includegraphics[width=0.4\columnwidth]{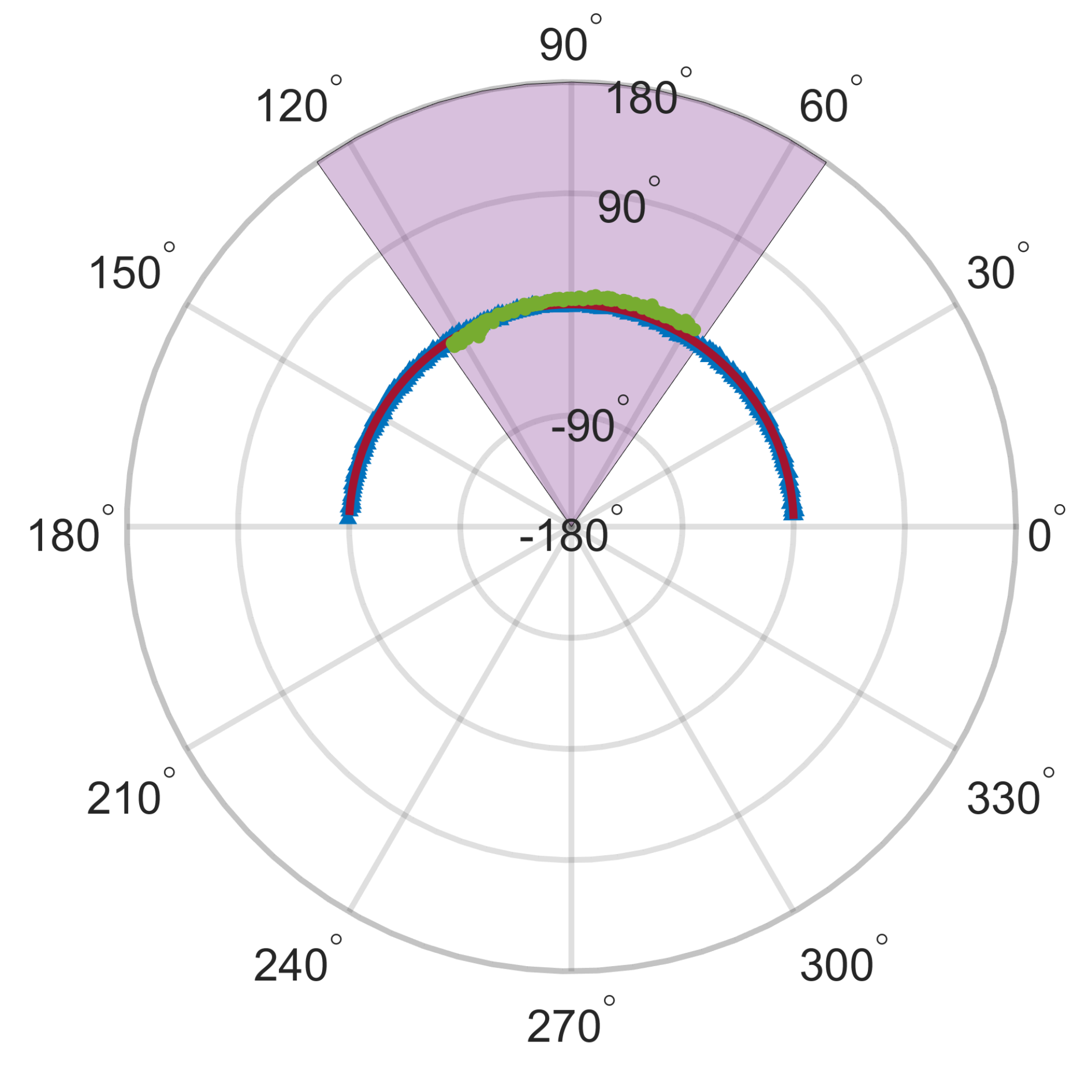}\\
             
             \cmidrule(lr){1-1}
            \cmidrule(lr){2-2}
            \cmidrule(lr){3-3}
            \multicolumn{1}{c}{ Yaw}&&\\
            
             &\includegraphics[width=0.4\columnwidth]{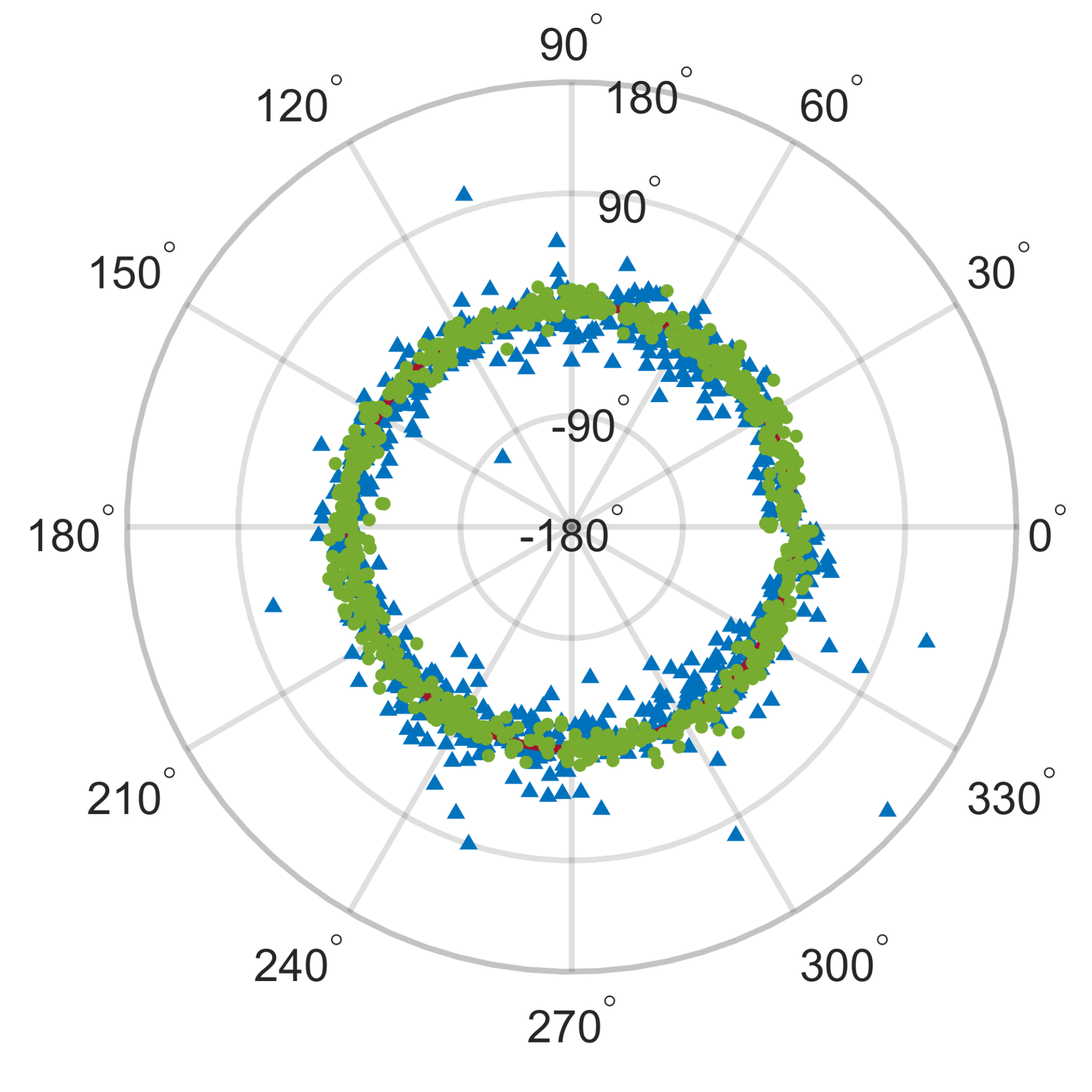} & \includegraphics[width=0.4\columnwidth]{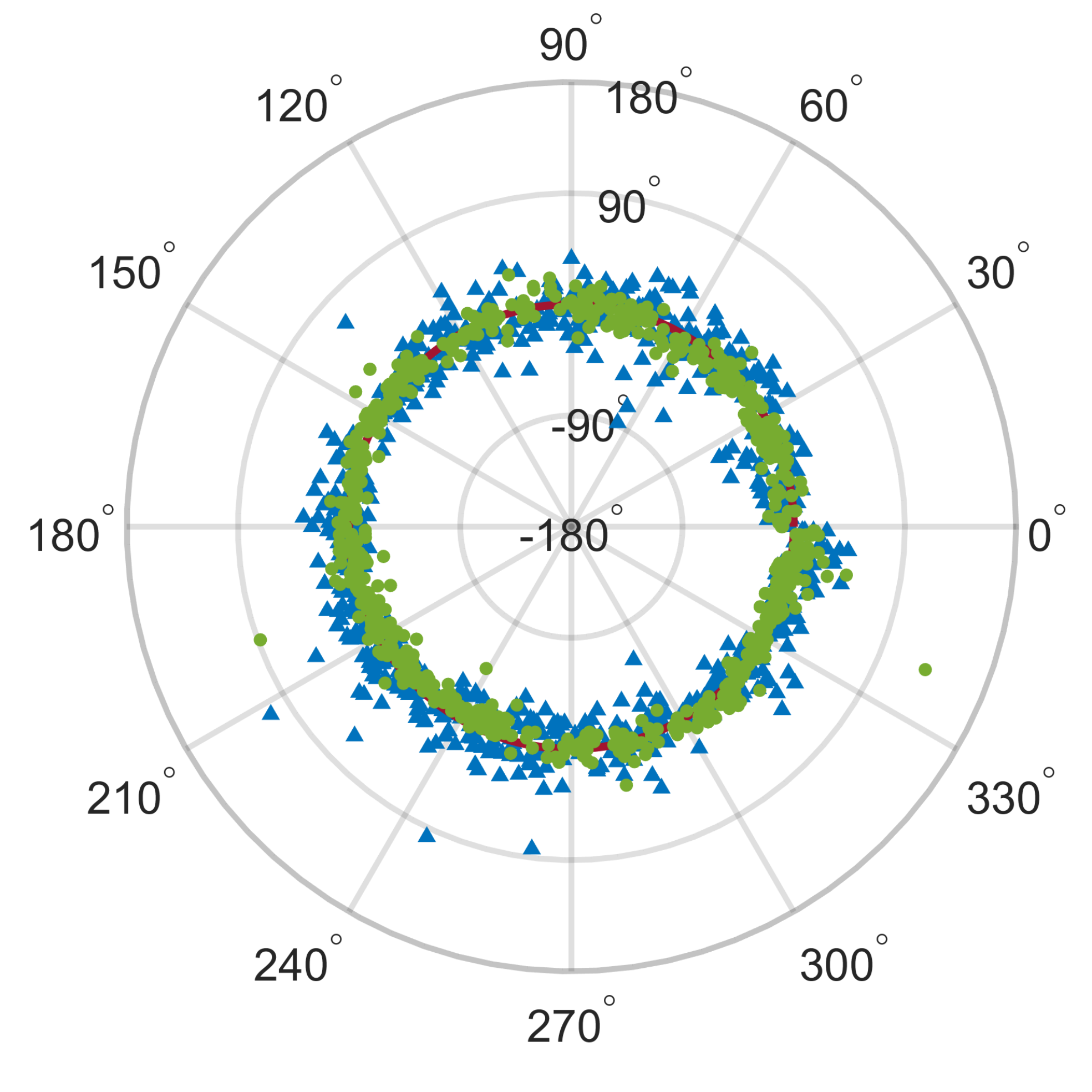}\\
             &\multicolumn{2}{c}{\includegraphics[width=0.65\columnwidth]{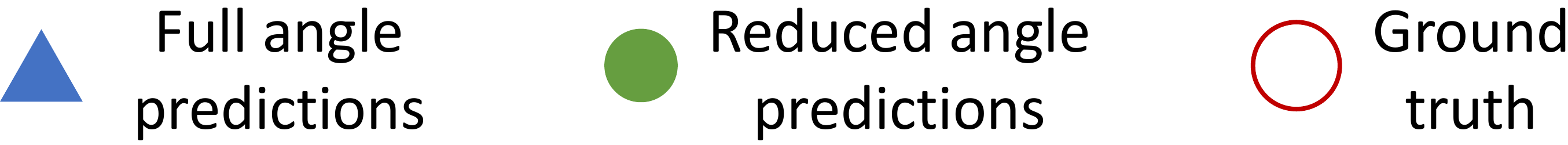}}
        \end{tabular}
        \captionof{figure}{The results of the orientation prediction networks for the two drones in the full angle and reduced angle regimes. The theta coordinate represents the angle with the solid red ring indicating the ground truth. The radial coordinate represents the error (up to a maximum of $\pm180^{\circ}$) with network under and over predictions falling inside of and outside of the red ring respectively. Predictions made by the network trained on the full range of angles are shown as blue triangles. Predictions made by the network trained on the reduced range of angles (indicated by the shaded region) are shown as green dots. }
	    \label{fig: Orientation_Hist_And_Inten}
    \end{figure}
\begin{figure}[t!]
	\centering
	\setlength\tabcolsep{3pt}
	\footnotesize
        \begin{tabular}{ c c c }
        
            \toprule
            \multicolumn{3}{c}{\bfseries Mavic 2}\\
            \cmidrule[\heavyrulewidth](lr){1-3}
            Truth&&Prediction\\
            \cmidrule(lr){1-1}
            \cmidrule(lr){3-3}
            \multicolumn{1}{c}{\includegraphics[width=0.4\columnwidth]{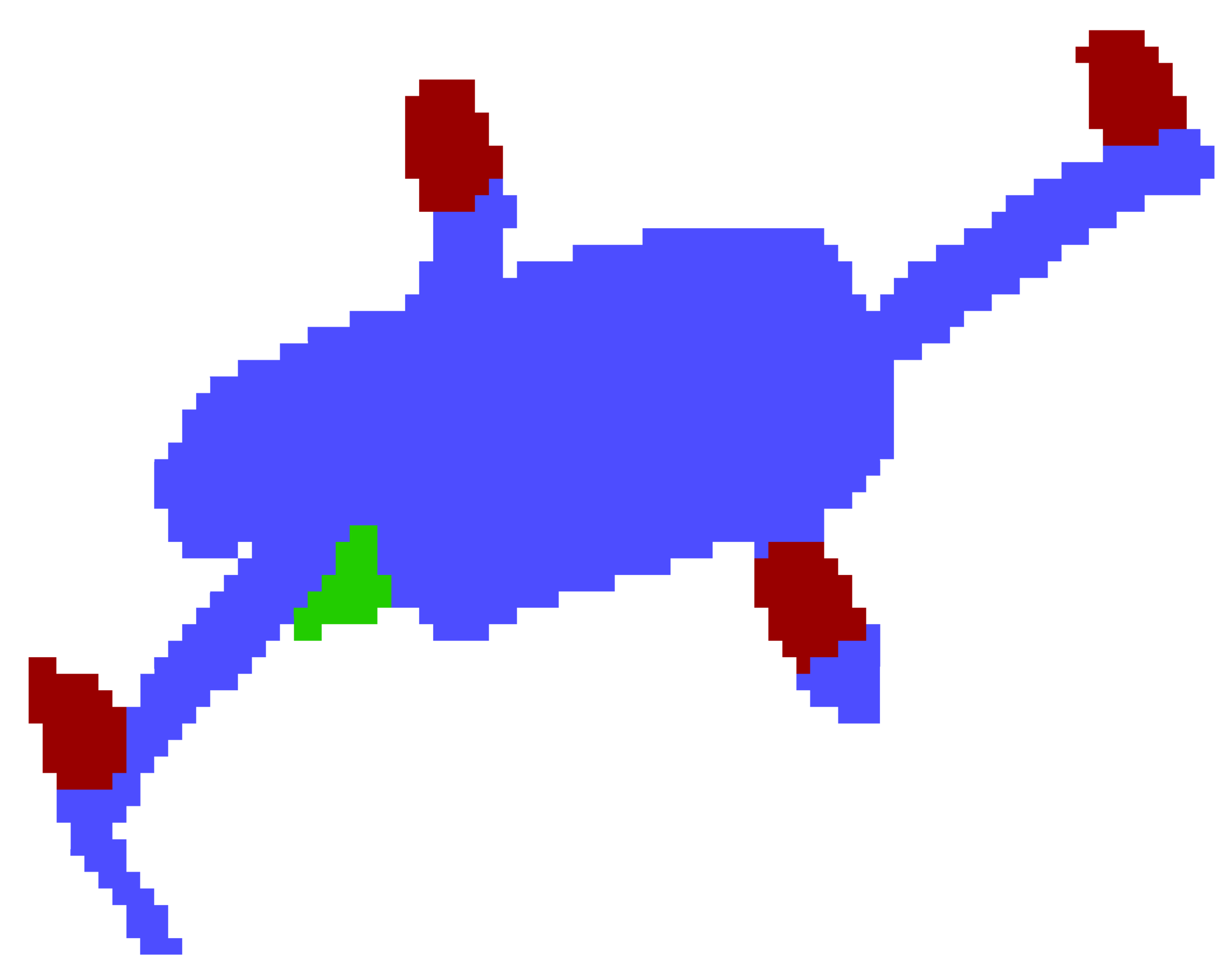}}  & \multicolumn{1}{c}{\includegraphics[width=0.1\columnwidth,height=1.5cm]{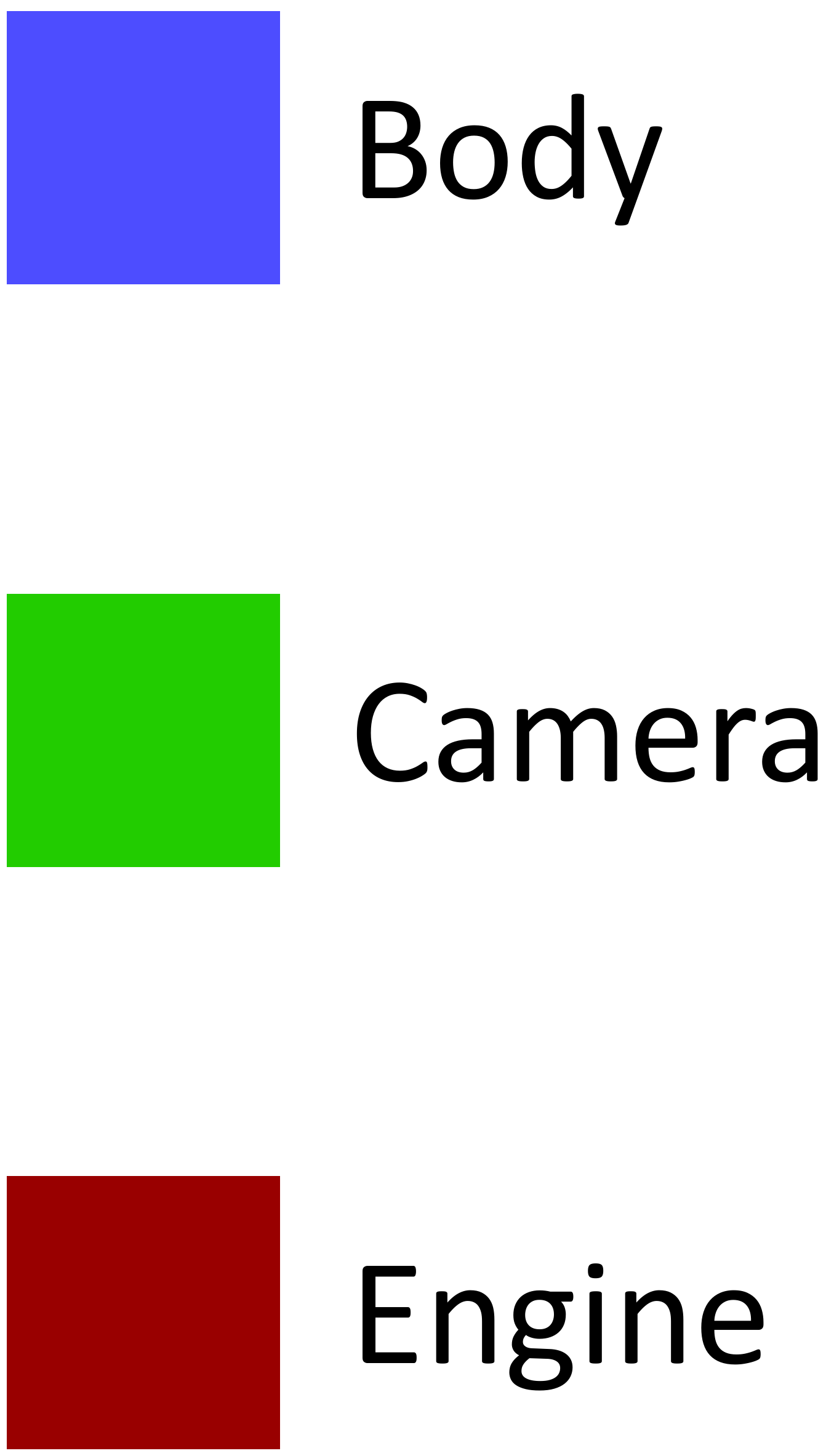}}
            &\multicolumn{1}{c}{\includegraphics[width=0.4\columnwidth]{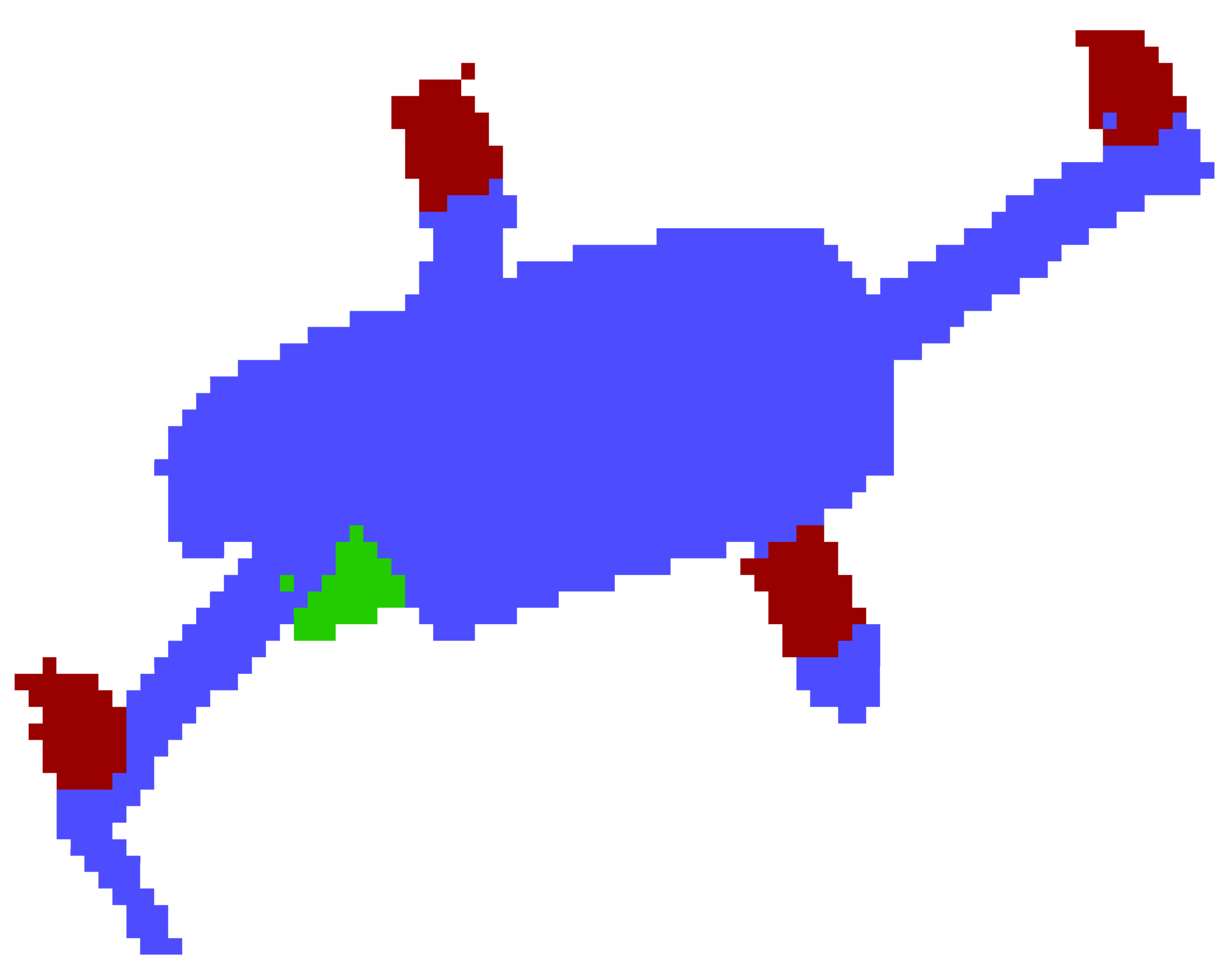}}\\
            \cmidrule(lr){1-3}
            \multicolumn{1}{c}{\adjincludegraphics[width=0.4\columnwidth,trim={0 0 {.15\width} 0},clip]{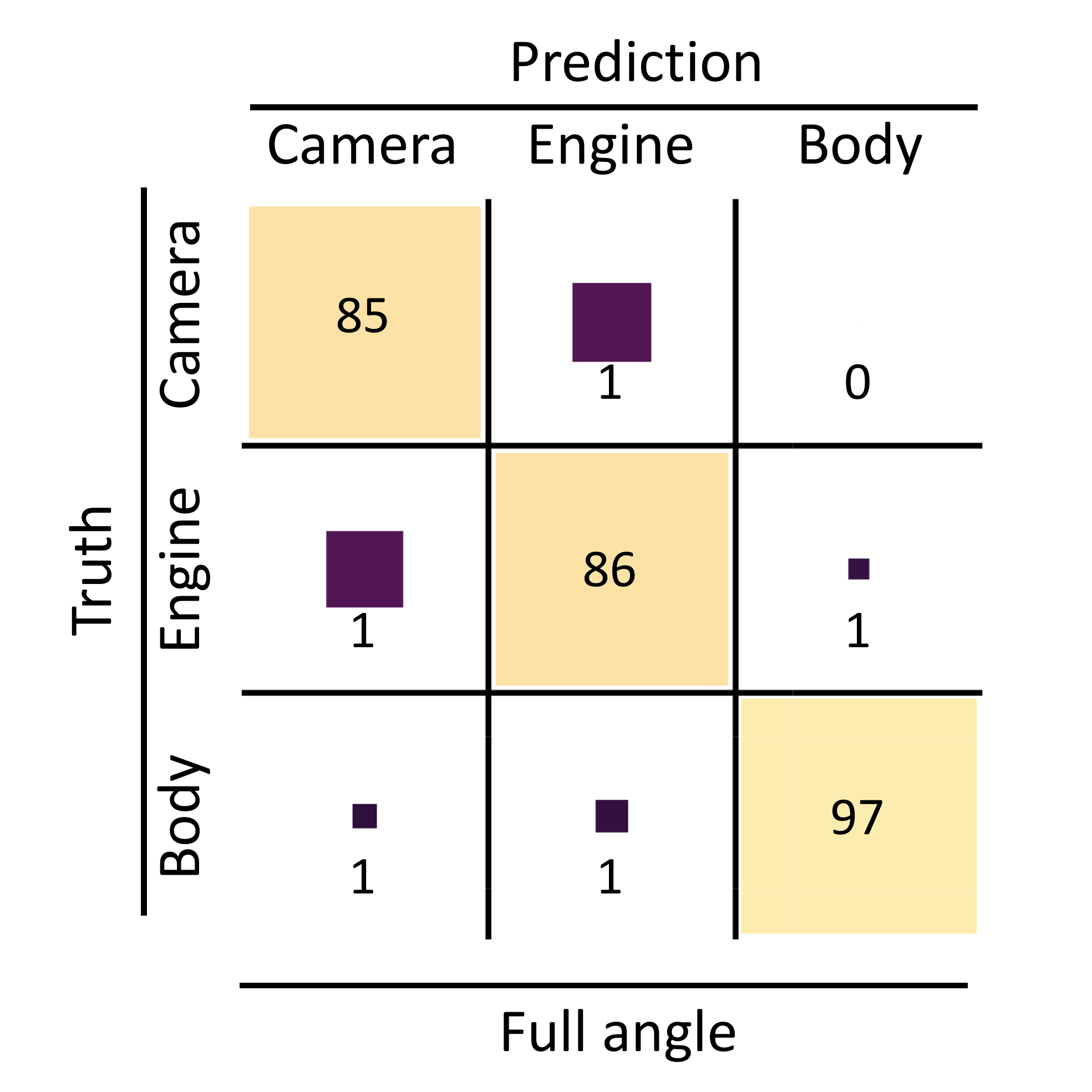}}&
            \multicolumn{1}{c}{\hspace{0.25cm}\includegraphics[width=0.06\columnwidth,height=3.6cm]{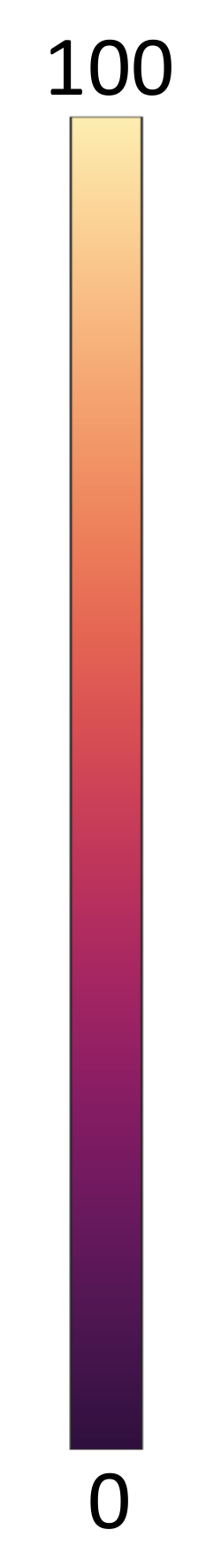}}
            &\multicolumn{1}{c}{\adjincludegraphics[width=0.4\columnwidth,trim={0 0 {.15\width} 0},clip]{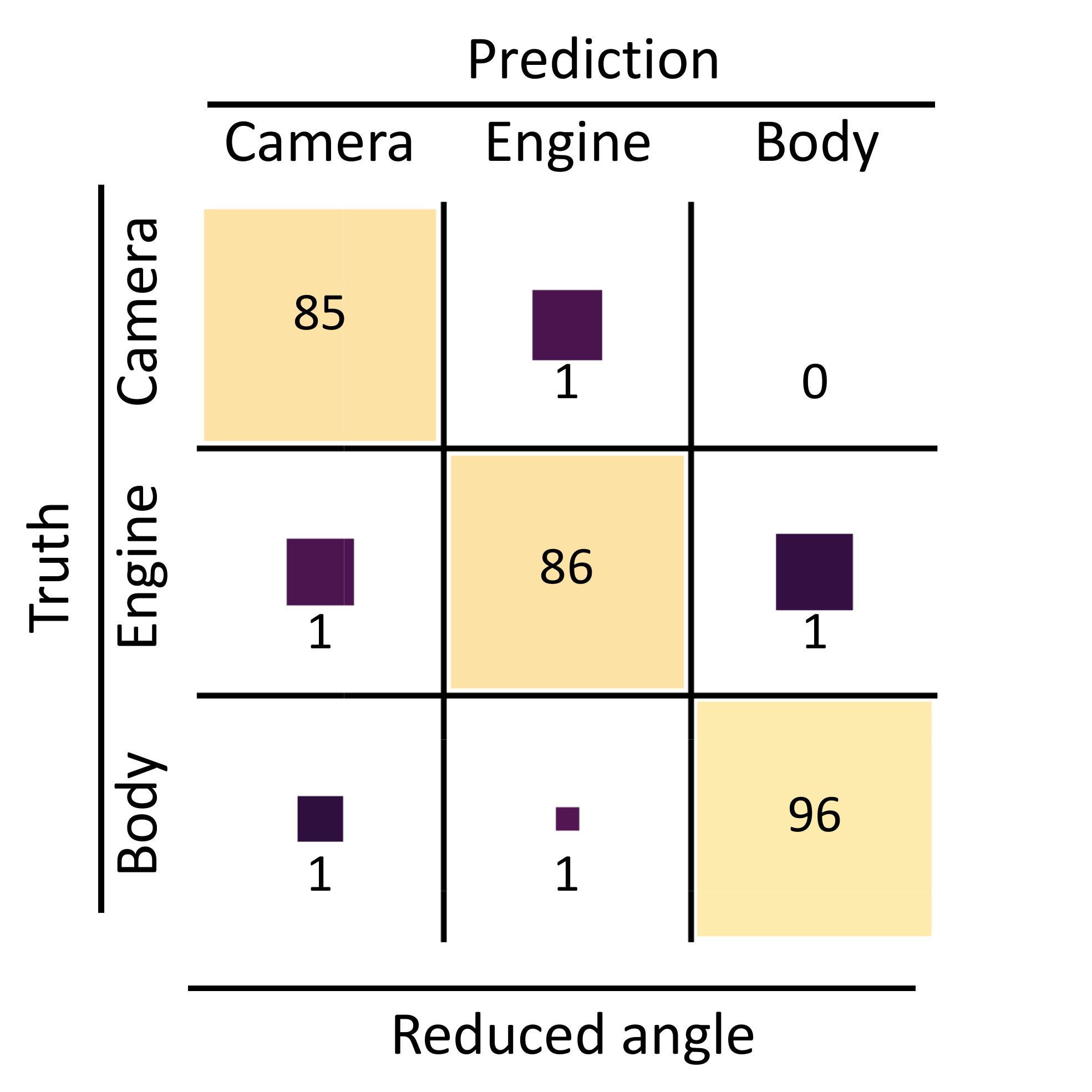}}\\
            \cmidrule[\heavyrulewidth](lr){1-3}
             \multicolumn{3}{c}{\bfseries Inspire 2}\\
            \cmidrule[\heavyrulewidth](lr){1-3}
            Truth&&Prediction\\
            \cmidrule(lr){1-1}
            \cmidrule(lr){3-3}
            
            \multicolumn{1}{c}{\includegraphics[width=0.4\columnwidth]{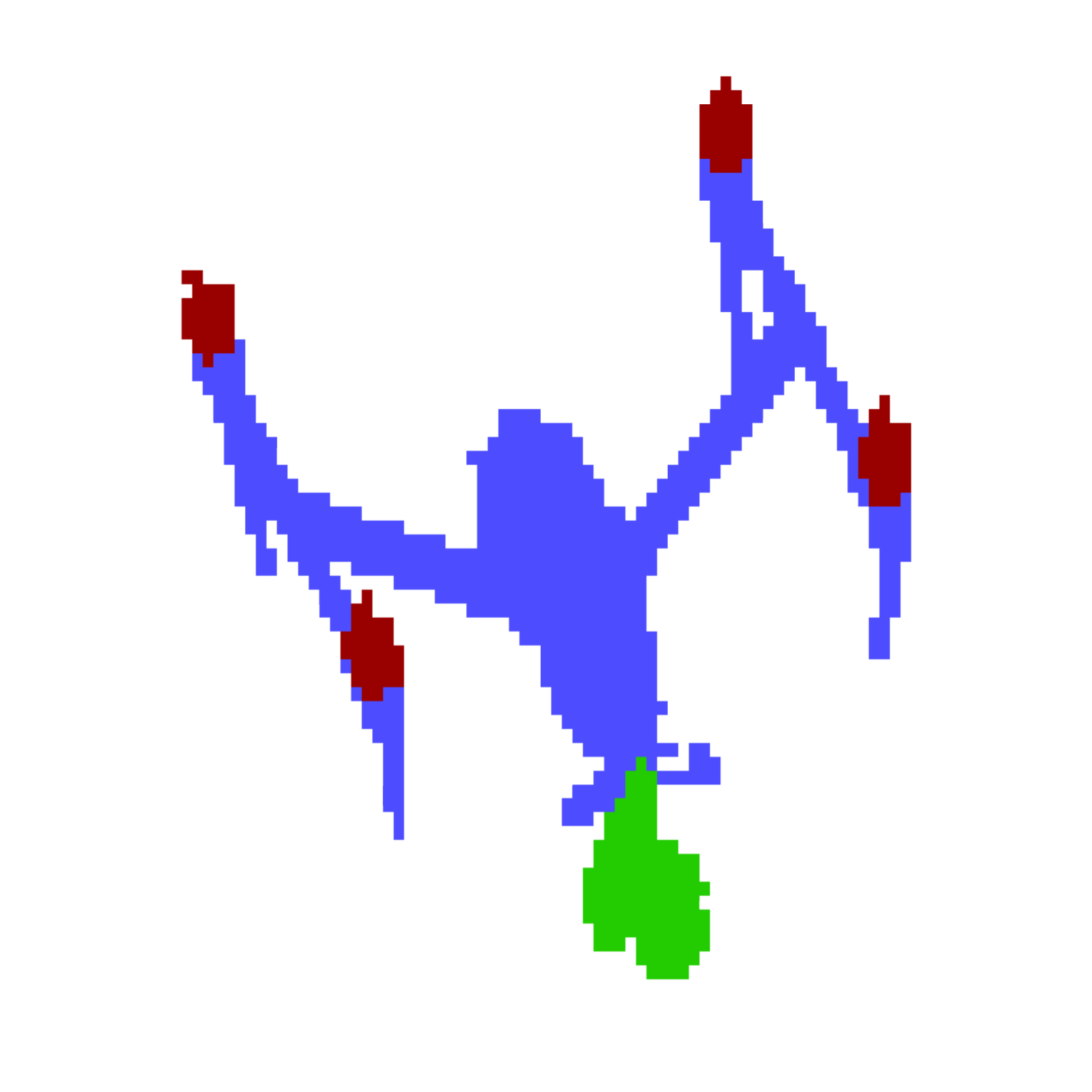}} 
            &\multicolumn{1}{c}{\includegraphics[width=0.1\columnwidth,height=1.5cm]{Graphics/Schol_23.pdf}}
            &\multicolumn{1}{c}{\includegraphics[width=0.4\columnwidth]{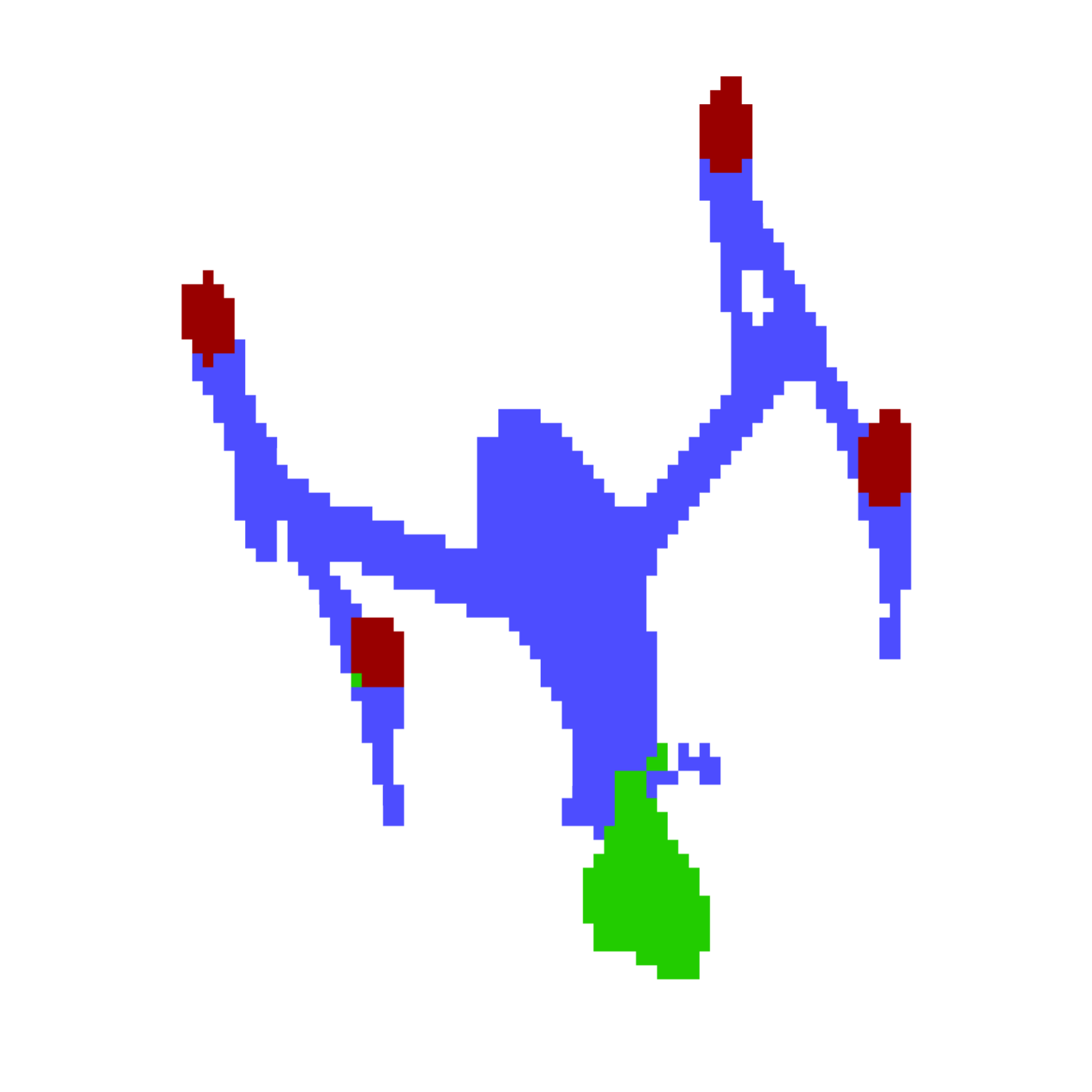}}\\
            \cmidrule(lr){1-3}
            
            
            \multicolumn{1}{c}{\adjincludegraphics[width=0.4\columnwidth,trim={0 0 {.15\width} 0},clip]{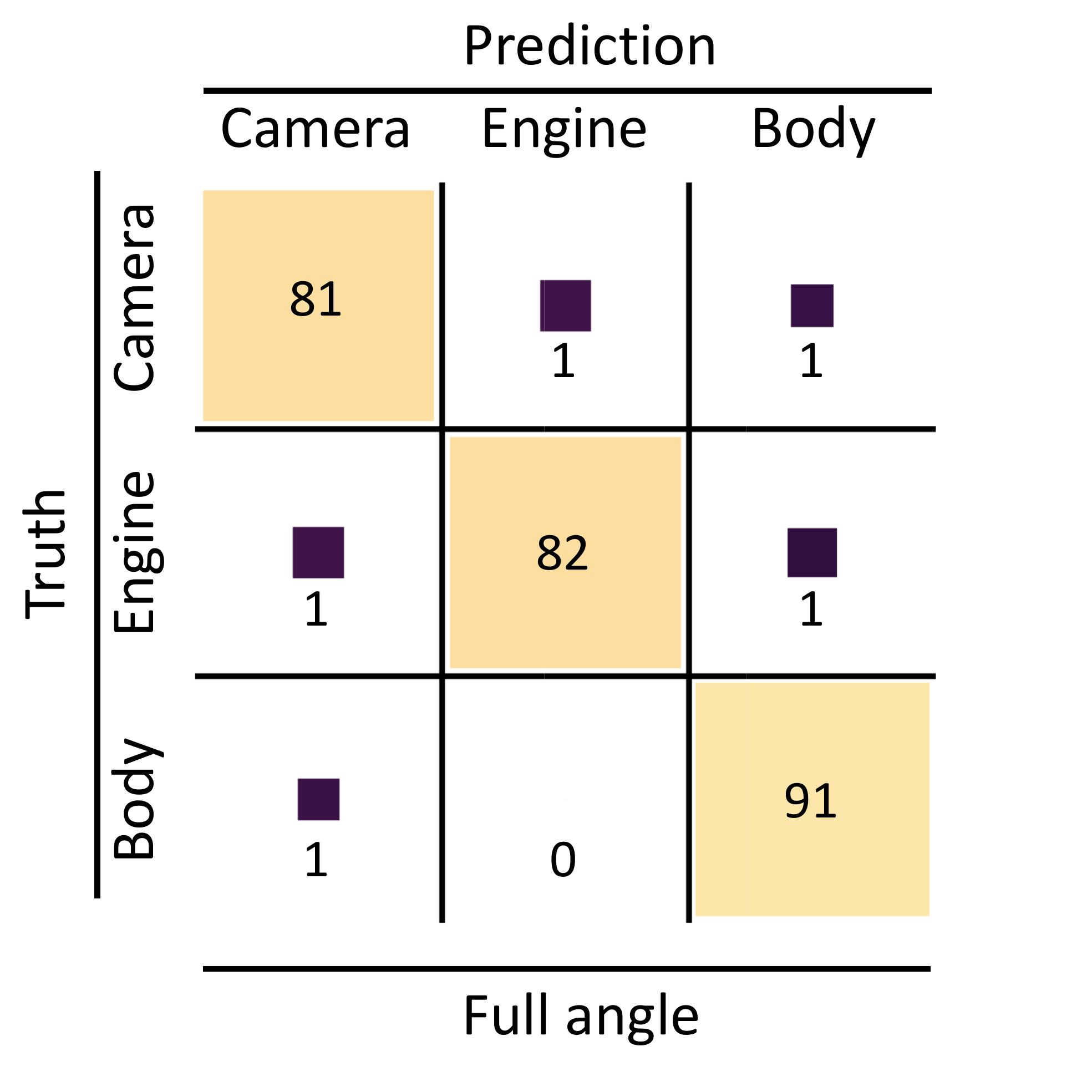}} &
            \multicolumn{1}{c}{\includegraphics[height=0.1\columnwidth,height=3.6cm]{Graphics/Schol_26.pdf}}&
            \multicolumn{1}{c}{\adjincludegraphics[width=0.4\columnwidth,trim={0 0 {.15\width} 0},clip]{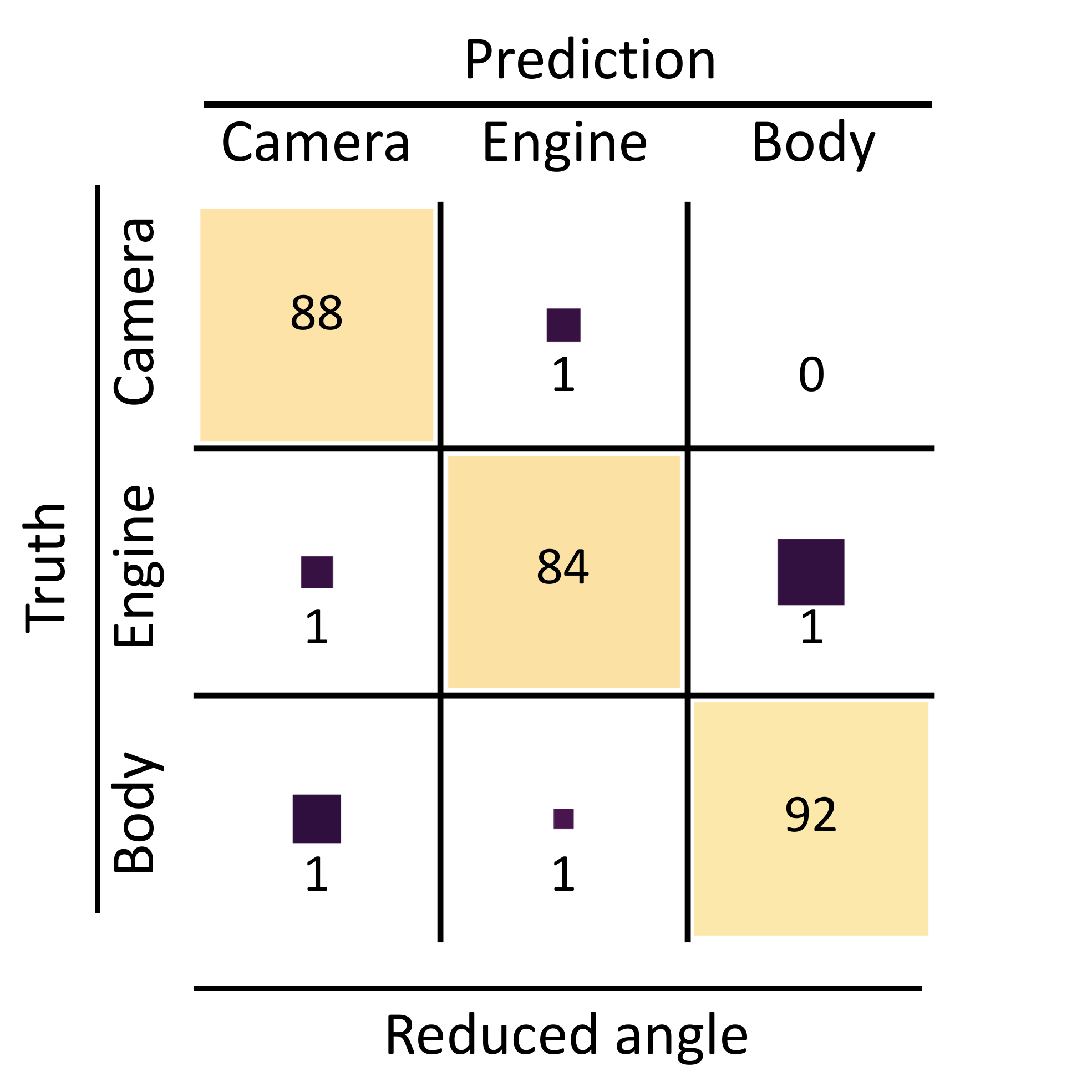}}\\
  
	        \bottomrule
        \end{tabular}
        \captionof{figure}{Qualitative and quantitative analyses of the segmentation networks. The quantitative analyses uses the IoU percentage for the two drones in the full angle and reduced angle regimes. The number in, as well as the size of each coloured region corresponds to the nearest integer IoU percentage. Note that the sizes of the regions have been scaled logarithmically for clearer representation.}
	    \label{fig: Segmentation}
\end{figure}
\noindent Fig.~\ref{fig: Orientation_Hist_And_Inten} displays the predictions of the orientation networks for the full angle and reduced angle regimes. The theta coordinate represents the angle and the radial coordinate represents the error with $-180^{\circ}$ error at the center and $+180^{\circ}$ error at the circumference. The solid red ring indicates the ground truth. Network under and over predictions fall inside of and outside of the red ring respectively. Predictions made by the network trained on the full range of angles are shown as blue triangles.  Predictions made by the network trained on the reduced range of angles (indicated by the shaded region) are shown as green dots.\\
\\
By examining the radial distribution of the predictions, the accuracy of the networks in each axis and in each regime can be compared and the following observations made. First, the accuracy of the networks is contingent upon the number of images-per-angle the network is given to train on. In the full angle regime where the pitch is restricted to half the range of the roll and yaw the network accuracy improves significantly since for the same number of total training images the number of examples-per-degree is doubled to $\sim$400. This is also why in the reduced angle regime where the roll is restricted its accuracy matches that of the pitch, while the yaw does not, even when the same total number of training images is used. Second, the accuracy of the networks is coupled i.e., for a reduced range of motion in one axis the accuracy of the remaining axes will increase. While this effect is less pronounced than that of examples-per-degree it can be observed in Table.~\ref{tab: Results_Summary} where a $4^{\circ}$ increase in yaw accuracy is observed for both drones in the reduced angle regime. This despite the range of motion in that axis remaining constant. The improvment can be attributed to the reduced variance (roll and pitch range) in the images which the yaw network must learn. Third, the accuracy of the networks is somewhat contingent on the symmetry of the drone. Specifically, the Mavic 2 is nearly perfectly symmetric about it's roll axis, consequently the accuracy of the Mavic 2 roll network in the full angle regime is the worst. This is because there are the fewest features to unambiguously identify the roll at angles outside of a $90^{\circ}\text{ to }270^{\circ}$ range.\\
\\
Examining the Intersection over Union (IoU) scores in Fig.~\ref{fig: Segmentation} it is apparent that the networks can effectively segment both drones into their components regardless of their orientation. The score relating to the `body' label is the highest in all cases indicating that the network is most accurate at predicting this component. This is likely because it is the most prevalent in terms of pixels in the image. Additionally, the fact that the rows and columns of the IoU scores do not sum to 100 indicates a conservative predictor. This means the network leaves some pixels (particularly around the perimeter of the drone) unclassified, reducing the total accuracy but also minimising misclassification.
\subsection{Reduced input results}
\label{subsec: Reduced input results}
\begin{figure}[t!]
	\centering
	    \footnotesize
        \begin{tabular}{ c c c  }
             \multicolumn{1}{c}{\bfseries   Axis}&\multicolumn{1}{c}{\bfseries   Intensity only}& \multicolumn{1}{c}{\bfseries   Histograms only}\\
             
             \cmidrule(lr){1-1}
            \cmidrule(lr){2-2}
            \cmidrule(lr){3-3}
            \multicolumn{1}{c}{ Roll}&&\\
              &\includegraphics[width=0.4\columnwidth]{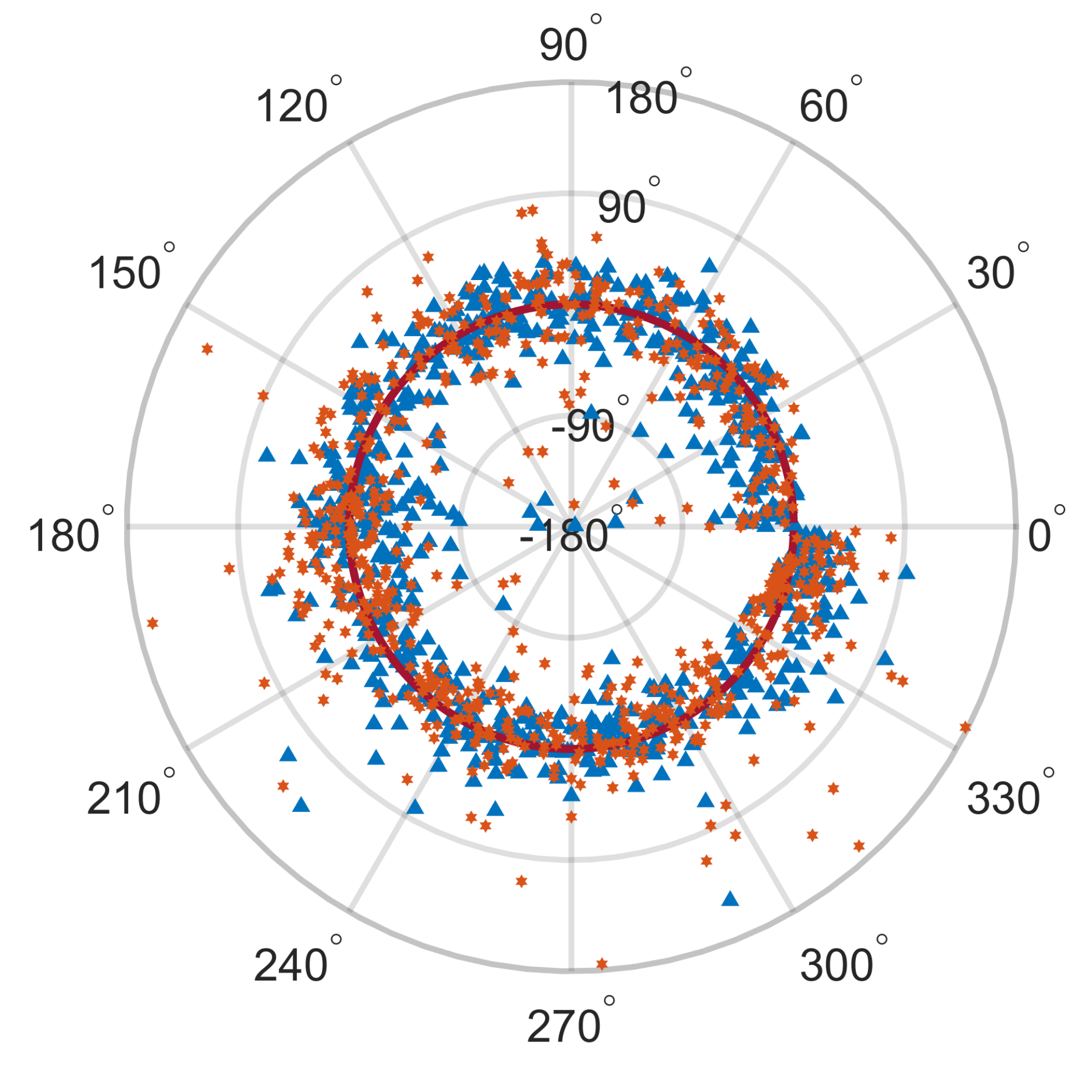} & \includegraphics[width=0.4\columnwidth]{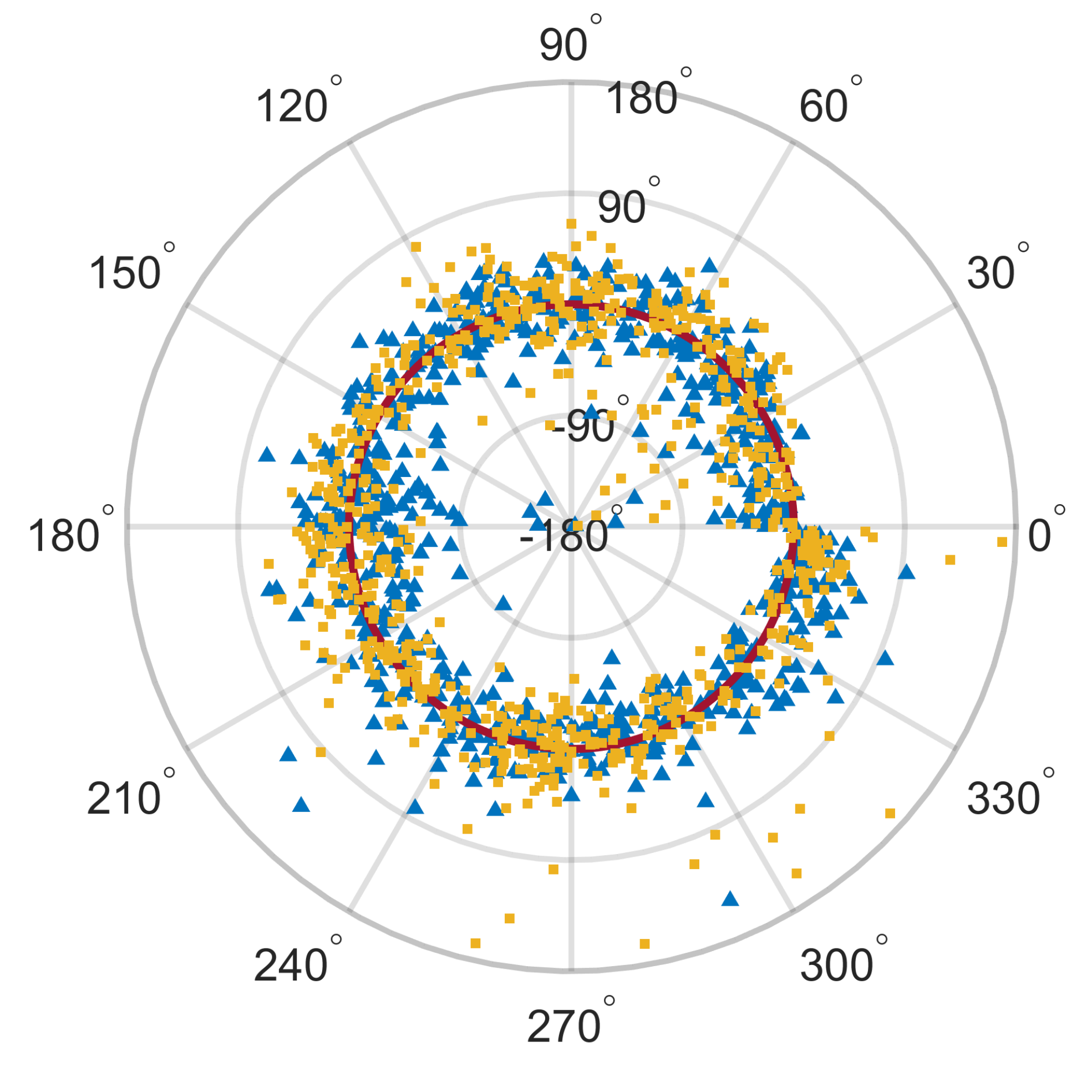}\\
             
             \cmidrule(lr){1-1}
            \cmidrule(lr){2-2}
            \cmidrule(lr){3-3}
            \multicolumn{1}{c}{ Pitch}&&\\
            
             &\includegraphics[width=0.4\columnwidth]{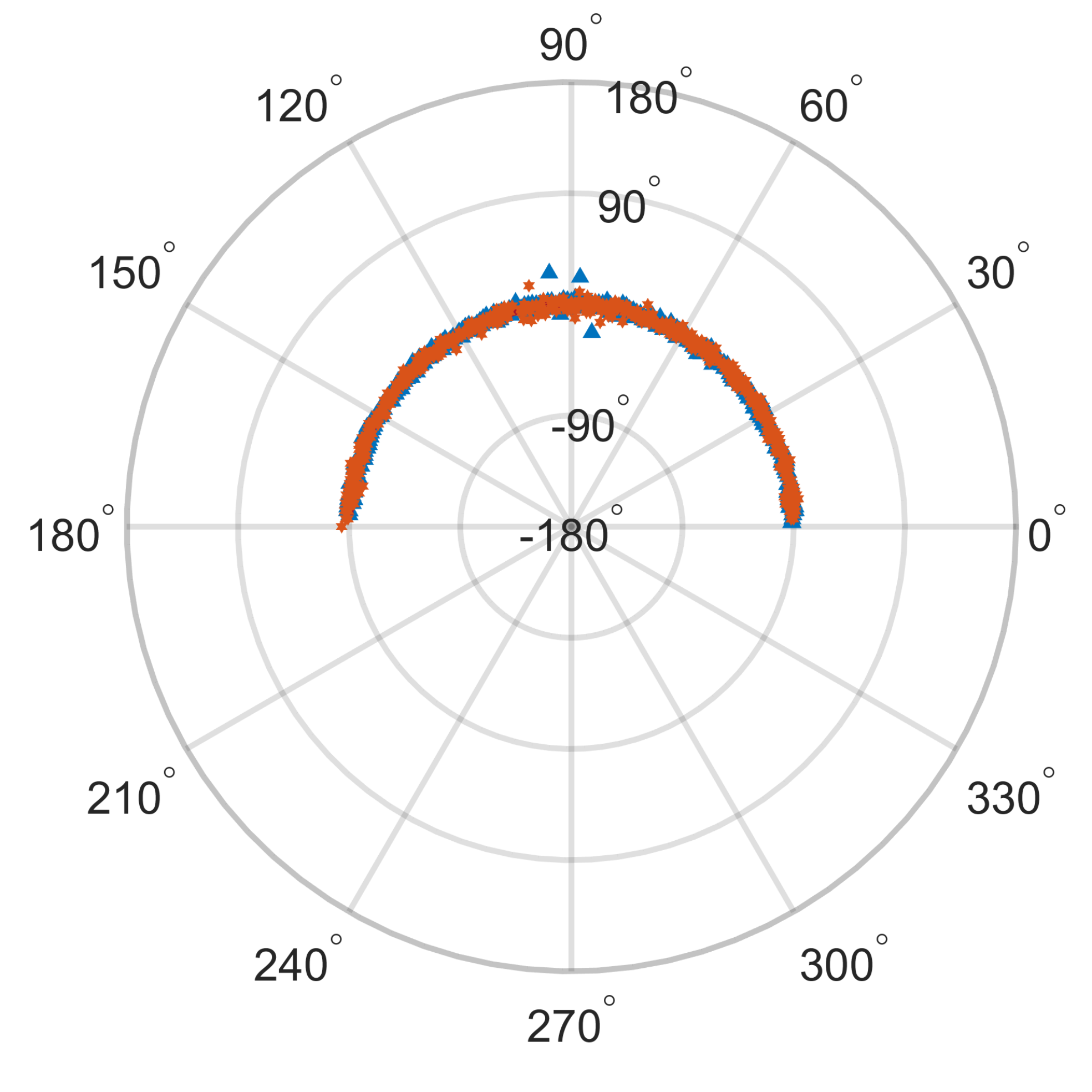} &
             \includegraphics[width=0.4\columnwidth]{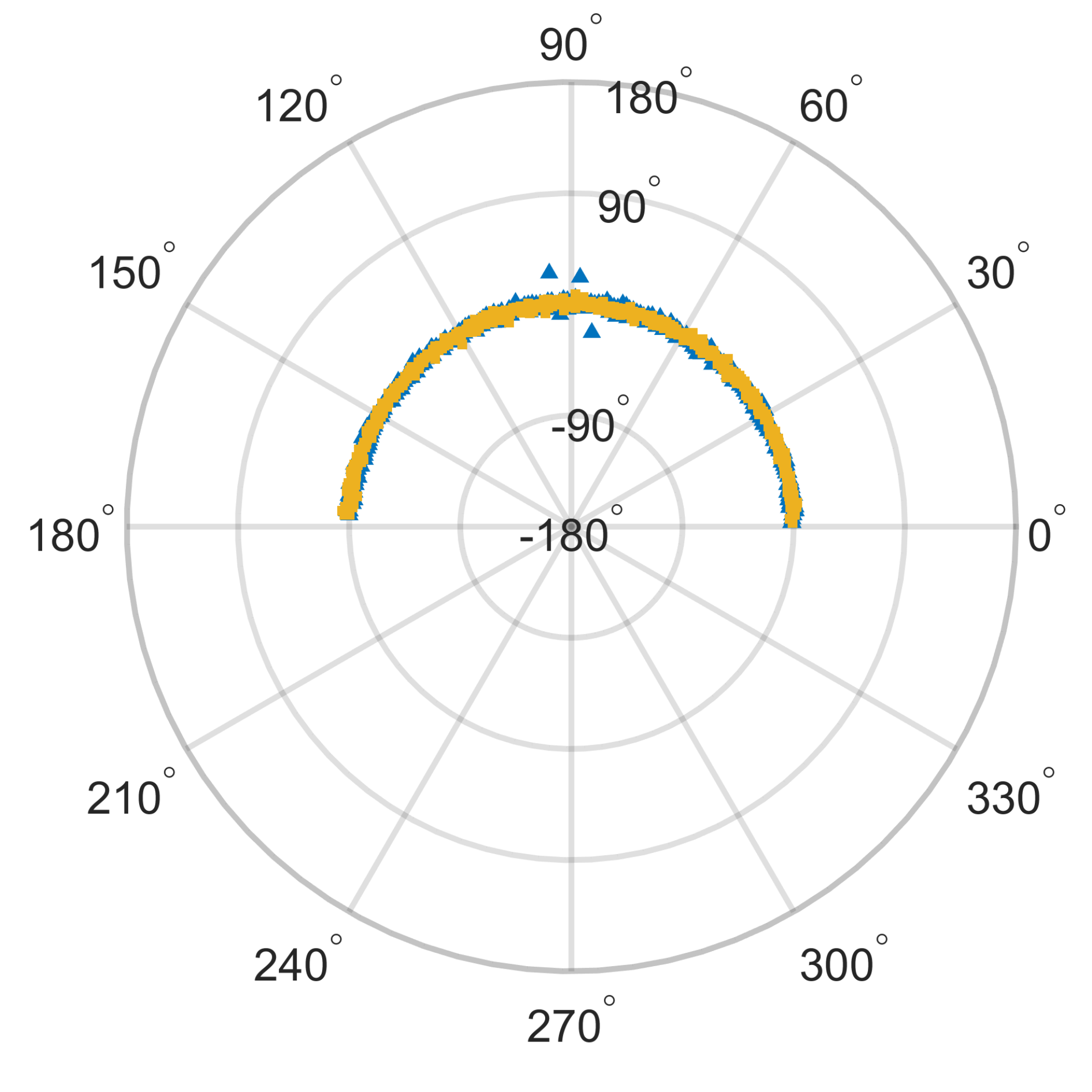}\\
             
             \cmidrule(lr){1-1}
            \cmidrule(lr){2-2}
            \cmidrule(lr){3-3}
            \multicolumn{1}{c}{ Yaw}&&\\
            
             &\includegraphics[width=0.4\columnwidth]{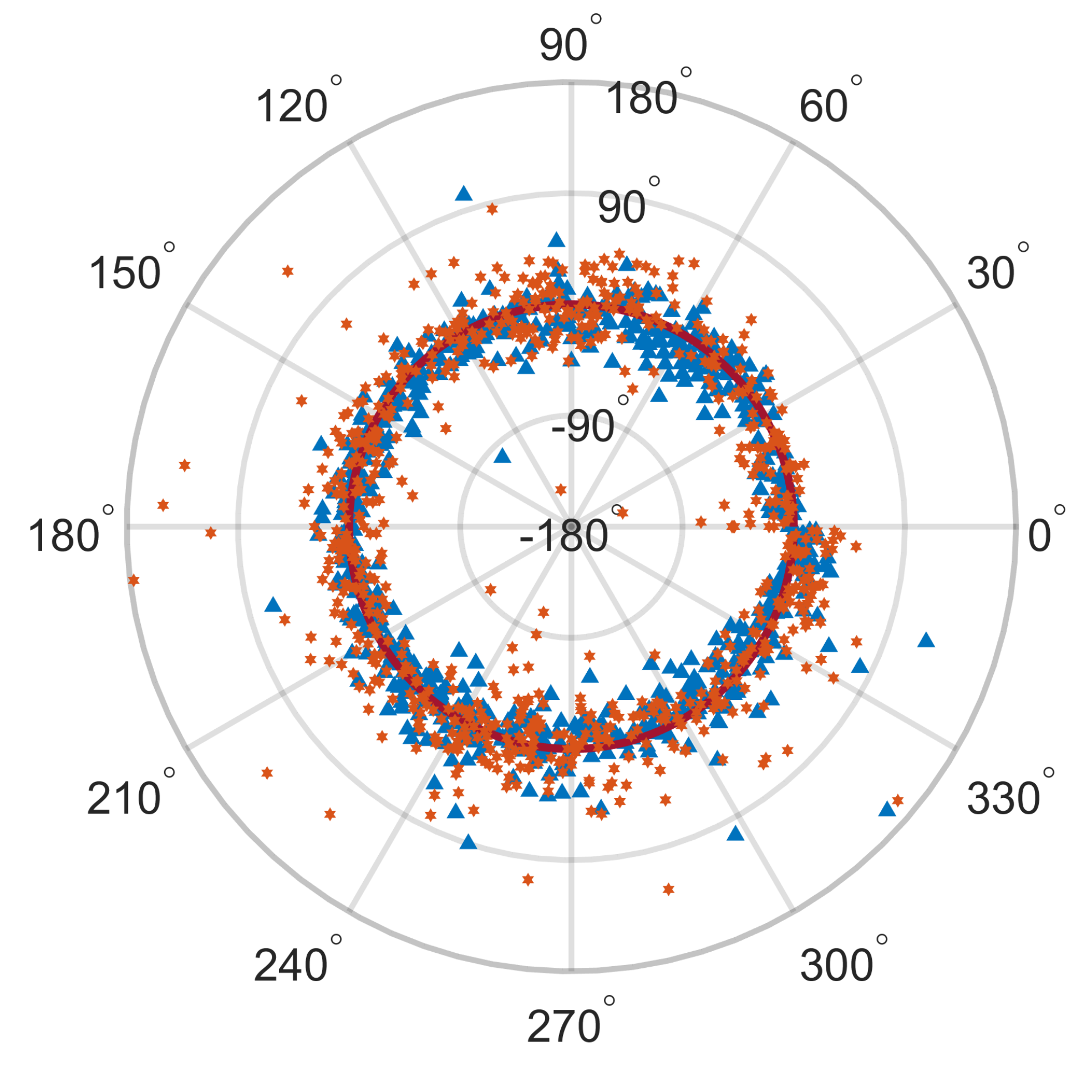} & \includegraphics[width=0.4\columnwidth]{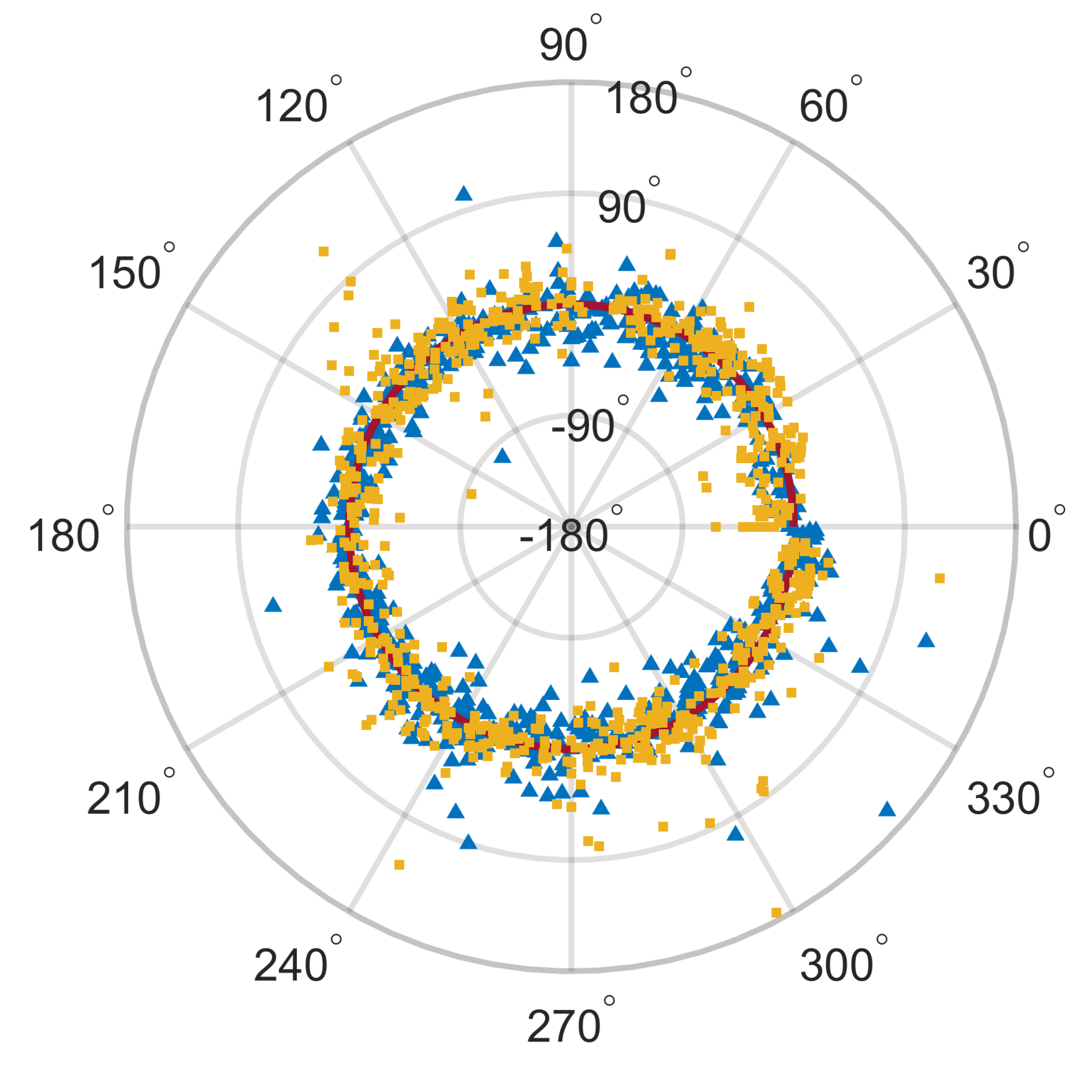}\\
             &\multicolumn{2}{c}{\includegraphics[width=0.8\columnwidth]{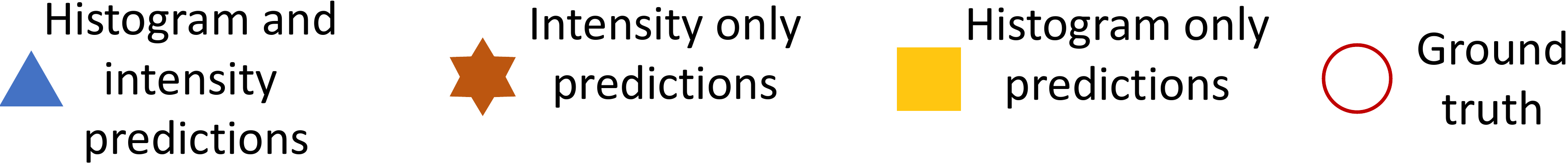}}
        \end{tabular}
        \captionof{figure}{The results of the orientation prediction networks for the Mavic 2 drone in the full angle regime when trained using only an intensity or depth input. The theta coordinate represents the angle with the solid red ring indicating the ground truth. The radial coordinate represents the error (up to a maximum of $\pm180^{\circ}$). Network under and over predictions fall inside of and outside of the red ring respectively. Predictions made by the networks trained on both inputs are shown as blue triangles. Predictions made by the networks trained on only intensity or depth data are shown as orange stars and yellow squares respectively. }
	    \label{fig: Orientation_Hist_Or_Inten}
    \end{figure}

To further examine the functioning of the networks an ablation study was conducted. Specifically, the effect of removing one input channel, either the histograms or the intensity was quantified. Given that all networks share the DFE it was determined to be sufficient to retrain only the orientation network for the Mavic 2 in the full angle regime since changes in performance in this network would be indicative of changes in all networks. Table~\ref{tab: Hist_Or_Inten_Results_Summary} presents a summary of the findings with the network predictions visualized in Fig.~\ref{fig: Orientation_Hist_Or_Inten}.
\begin{table}[t!]
    \caption{Summary of the Mavic 2's orientation network accuracy when trained using only an intensity input or a depth input in the full angle regime.}
	\centering
	\footnotesize
        \begin{tabular}{ c c c}
            \toprule
           \textbf{  Metric} & \textbf{ Histogram only} & \textbf{ Change}\\
            
            \cmidrule[\heavyrulewidth](lr){1-1}
            \cmidrule[\heavyrulewidth](lr){2-2}
            \cmidrule[\heavyrulewidth](lr){3-3}
            Orientation & (accuracy $\pm$ std)(\%)  & (accuracy ; std)(\%) \\
	        \cmidrule(lr){1-1}

	        Roll &  87.4$\pm$13.7& -0.9 ; +0.7\\ 
	        \cmidrule(lr){2-2}
	        \cmidrule(lr){3-3}

	        Pitch &  99.1$\pm$0.9& -0.1 ; +0.1 \\
	        \cmidrule(lr){2-2}
	        \cmidrule(lr){3-3}
	        
	        Yaw &  91.9$\pm$10.1& -0.4 ; +0.4 \\
	        
	        \cmidrule[\heavyrulewidth](lr){1-3}
	        & \textbf{ Intensity only} & \textbf{ Change}\\
	        \cmidrule[\heavyrulewidth](lr){2-2}
            \cmidrule[\heavyrulewidth](lr){3-3}
	        Orientation & (accuracy $\pm$ std)(\%)  & (accuracy ; std)(\%) \\
	        \cmidrule(lr){1-1}

	        Roll  & 86.6$\pm$14.2& -1.7 ; +1.2\\
	        \cmidrule(lr){2-2}
	        \cmidrule(lr){3-3}

	        Pitch  & 98.8$\pm$1.2& -0.4 ; +0.4\\
            \cmidrule(lr){2-2}
	        \cmidrule(lr){3-3}
	        
	        Yaw  &88.6$\pm$13.1& -3.6 ; +3.4\\

	        \bottomrule
        \end{tabular}
	    \label{tab: Hist_Or_Inten_Results_Summary}
    \end{table}
Table~\ref{tab: Hist_Or_Inten_Results_Summary} and Fig.~\ref{fig: Orientation_Hist_Or_Inten} indicate that the orientation of a drone can be more accurately determined from a depth input than an intensity input although the relative improvement is small. It should be noted however, that the images on which the network was trained do not contain a background. In real world cases where drones could be optically camouflaged the ability for depth sensing devices to isolate volumes of space ahead of background objects using time-of-flight gating may significantly enhance their robustness in orientation detection. Additionally, given that the segmentation network can only reliably produce images up to the size of it's largest input (due to its U-Net structure) there is a benefit to providing the network with a high transverse resolution image.
\subsection{Results on real data}
\label{subsec:results on real data}
\begin{figure}[b!]
	\centering
	    \setlength\tabcolsep{1pt}
	    \footnotesize
        \begin{tabular}{  c c c  }
            \toprule
             
            \textbf{ Metric} & \textbf{Ground truth} & \textbf{Prediction; Accuracy (\%) } \\
            
            \cmidrule[\heavyrulewidth](lr){1-1}
            \cmidrule[\heavyrulewidth](lr){2-2}
            \cmidrule[\heavyrulewidth](lr){3-3}
            Identification &  Mavic 2 & Mavic 2; 100\\
             \cmidrule(lr){1-1}
             \cmidrule(lr){2-2}
             \cmidrule(lr){3-3}
            Orientation &  & \\
             
	        \cmidrule(lr){1-1}
            Yaw &   31$^{\circ}$&19$^{\circ}$; 93 \\
	        \cmidrule(lr){2-2}
	        \cmidrule(lr){3-3}

	         Roll & 180$^{\circ}$&  177$^{\circ} $; 96 \\
	        \cmidrule(lr){2-2}
	        \cmidrule(lr){3-3}
	 
	        Pitch  & 90$^{\circ}$&   92 $^{\circ}$; 96 \\
	        \cmidrule[\heavyrulewidth](lr){1-3}
	        \textbf{ Metric} & \textbf{Reference} & \textbf{Prediction } \\
            
            \cmidrule[\heavyrulewidth](lr){1-1}
            \cmidrule[\heavyrulewidth](lr){2-2}
            \cmidrule[\heavyrulewidth](lr){3-3}
            Segmentation &   &   \\
	        \cmidrule(lr){1-1}
            &\includegraphics[width=0.3\columnwidth]{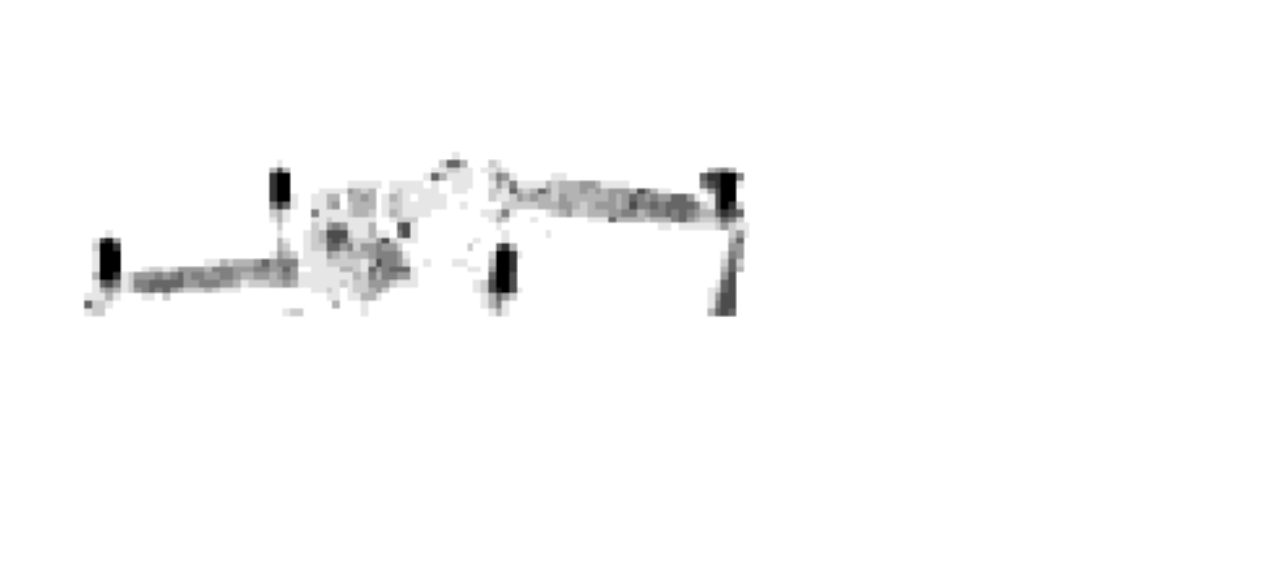} &
            \includegraphics[width=0.3\columnwidth]{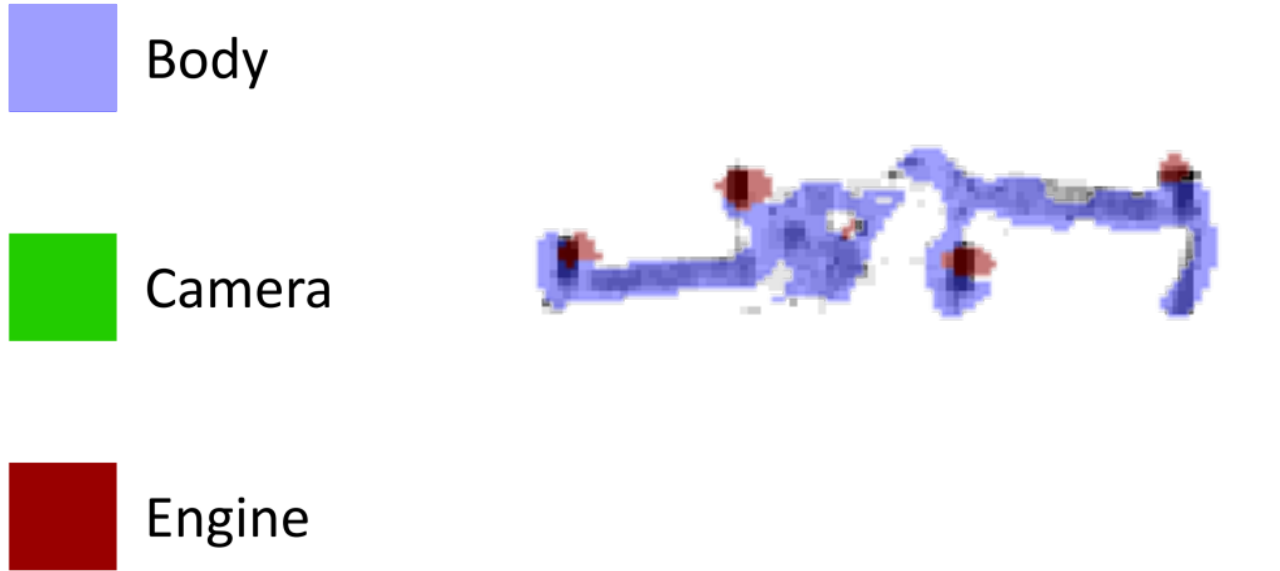} \\

	        \bottomrule
        \end{tabular}
        \captionof{figure}{The predictions of the network trained on simulated data when applied to Quantic 4x4 data of a real drone in flight. The network correctly predicted the drone type and identified the roll, pitch and yaw with an average accuracy of 95\%. The segmentation panels show the networks component prediction overlayed onto the intensity image (with the contrast enhanced intensity image shown alongside for reference). The network accurately segmented the drone into the body and engine components whilst not erroneously identifying a camera. }
	    \label{fig:Q4x4_Results}
    \end{figure}
To demonstrate the real world applicability of our system, we applied the reduced angle network (trained only on simulated data) to an image of a real DJI Mavic 2 Zoom drone captured in flight using a Quantic 4x4 SPAD camera. Fig.~\ref{fig:Q4x4_Results} summarizes the predictions made by the networks and highlights their ability to fully characterise drones in real world conditions. The network correctly identified the drone type and suffered only a small loss in accuracy when performing the segmentation and orientation operations. This reduction in accuracy can be attributed to the reduction in quality between the simulated data and the input data from the Quantic4x4 (as seen in Fig.~\ref{fig: Network inputs} c))

\section{Conclusion}
We present a CNN using a decision tree and ensemble structure to fully characterise i.e., determine the type, orientation and segmentation of drones in flight with accuracies in excess of 90\%. We provide a system for the rapid generation of large quantities of accurately labelled photo-realistic training data and demonstrate that this data is of sufficient fidelity to allow the system to accurately characterise real drones in flight. Our network provides a valuable tool in the image processing chain and can be used in combination with existing drone detection technologies to provide complete drone characterisation over wide areas. Finally, our approach may be readily extended to multiple 3D imaging and sensor fusion systems enabling pose detection for a wide range of vehicles. 

\section*{Acknowledgements}
The authors gratefully acknowledge the funding support of the Defence Science Technologies Laboratory through project Dstlx-1000147352. This work was supported by EPSRC through grants EP/T00097X/1 and EP/S026428/1.

\bibliographystyle{ieeetr}
\bibliography{mybibfile.bib}

\begin{thebibliography}{100}

\bibitem{gettinger2015drone}
D.~Gettinger and A.~H. Michel, ``Drone sightings and close encounters: An
  analysis,'' {\em Center for the Study of the Drone, Bard College,
  Annandale-on-Hudson, NY, USA}, 2015.

\bibitem{mehta2020before}
A.~M. Mehta, L.~Tam, D.~A. Greer, and K.~Letheren, ``Before crisis: How
  near-miss affects organizational trust and industry transference in emerging
  industries,'' {\em Public Relations Review}, vol.~46, no.~2, p.~101886, 2020.

\bibitem{rossiter2018drone}
A.~Rossiter, ``Drone usage by militant groups: exploring variation in
  adoption,'' {\em Defense \& Security Analysis}, vol.~34, no.~2, pp.~113--126,
  2018.

\bibitem{hammes2016democratization}
T.~Hammes, ``The democratization of airpower: The insurgent and the drone,''
  {\em War on the Rocks}, vol.~18, 2016.

\bibitem{schneider2019regulators}
D.~Schneider, ``Regulators seek ways to down rogue drones: Growing antidrone
  industry offers radar, remote id, and other tools-[news],'' {\em IEEE
  Spectrum}, vol.~56, no.~4, pp.~10--11, 2019.

\bibitem{nguyen2019towards}
P.~Nguyen, T.~Kim, J.~Miao, D.~Hesselius, E.~Kenneally, D.~Massey, E.~Frew,
  R.~Han, and T.~Vu, ``Towards rf-based localization of a drone and its
  controller,'' in {\em Proceedings of the 5th Workshop on Micro Aerial Vehicle
  Networks, Systems, and Applications}, pp.~21--26, 2019.

\bibitem{o2019no}
J.~O'Malley, ``The no drone zone,'' {\em Engineering \& Technology}, vol.~14,
  no.~2, pp.~34--38, 2019.

\bibitem{RN229}
I.~Bisio, C.~Garibotto, F.~Lavagetto, A.~Sciarrone, and S.~Zappatore,
  ``Unauthorized amateur uav detection based on wifi statistical fingerprint
  analysis,'' {\em Ieee Communications Magazine}, vol.~56, no.~4, pp.~106--111,
  2018.

\bibitem{RN230}
I.~Bisio, C.~Garibotto, F.~Lavagetto, A.~Sciarrone, S.~Zappatore, and Ieee,
  {\em Improving WiFi Statistical Fingerprint-based Detection Techniques
  against UAV Stealth Attacks}.
\newblock IEEE Global Communications Conference, New York: Ieee, 2018.

\bibitem{RN58}
H.~Zhang, C.~H. Cao, L.~W. Xu, and T.~A. Gulliver, ``A uav detection algorithm
  based on an artificial neural network,'' {\em Ieee Access}, vol.~6,
  pp.~24720--24728, 2018.

\bibitem{RN55}
M.~Ezuma, F.~Erden, C.~K. Anjinappa, O.~Ozdemir, I.~Guvenc, and Ieee, {\em
  Micro-UAV Detection and Classification from RF Fingerprints Using Machine
  Learning Techniques}.
\newblock IEEE Aerospace Conference Proceedings, New York: Ieee, 2019.

\bibitem{RN246}
M.~S. Allahham, T.~Khattab, and A.~Mohamed, ``Deep learning for rf-based drone
  detection and identification: A multi-channel 1-d convolutional neural
  networks approach,'' in {\em 2020 IEEE International Conference on
  Informatics, IoT, and Enabling Technologies (ICIoT)}, pp.~112--117, IEEE.

\bibitem{RN348}
D.~Shorten, A.~Williamson, S.~Srivastava, J.~C. Murray, and M.~Assoc~Comp, {\em
  Localisation of Drone Controllers from RF Signals using a Deep Learning
  Approach}.
\newblock Prai 2018: Proceedings of the International Conference on Pattern
  Recognition and Artificial Intelligence, New York: Assoc Computing Machinery,
  2018.

\bibitem{RN249}
S.~Al-Emadi and F.~Al-Senaid, ``Drone detection approach based on
  radio-frequency using convolutional neural network,'' in {\em 2020 IEEE
  International Conference on Informatics, IoT, and Enabling Technologies
  (ICIoT)}, pp.~29--34, IEEE.

\bibitem{RN168}
H.~C. Chen, Z.~Wang, and L.~Y. Zhang, ``Collaborative spectrum sensing for
  illegal drone detection: A deep learning-based image classification
  perspective,'' {\em China Communications}, vol.~17, no.~2, pp.~81--92, 2020.

\bibitem{RN171}
M.~U. Sheikh, F.~Ghavimi, K.~Ruttik, R.~Jantti, and Ieee, {\em Drone Detection
  and Classification Using Cellular Network: A Machine Learning Approach}.
\newblock IEEE Vehicular Technology Conference Proceedings, New York: Ieee,
  2019.

\bibitem{RN93}
A.~Bello, B.~Biswal, S.~Shetty, C.~Kamhoua, and K.~Gold, {\em Radio Frequency
  Classification Toolbox for Drone Detection}, vol.~11006 of {\em Proceedings
  of SPIE}.
\newblock Bellingham: Spie-Int Soc Optical Engineering, 2019.

\bibitem{RN203}
H.~Ryden, S.~Bin~Redhwan, X.~Q. Lin, and Ieee, {\em Rogue Drone Detection: A
  Machine Learning Approach}.
\newblock IEEE Wireless Communications and Networking Conference, New York:
  Ieee, 2019.

\bibitem{RN327}
A.~Shoufan, H.~M. Al-Angari, M.~F.~A. Sheikh, and E.~Damiani, ``Drone pilot
  identification by classifying radio-control signals,'' {\em Ieee Transactions
  on Information Forensics and Security}, vol.~13, no.~10, pp.~2439--2447,
  2018.

\bibitem{RN113}
D.~Mototolea, R.~Youssef, E.~Radoi, and I.~Nicolaescu, ``Non-cooperative
  low-complexity detection approach for fhss-gfsk drone control signals,'' {\em
  IEEE Open Journal of the Communications Society}, vol.~1, pp.~401--412, 2020.

\bibitem{RN227}
S.~Basak, S.~Rajendran, S.~Pollin, B.~Scheers, and Ieee, {\em Drone
  classification from RF fingerprints using deep residual nets}, pp.~548--555.
\newblock International Conference on Communication Systems and Networks, 2021.

\bibitem{RN259}
A.~Gumaei, M.~Al-Rakhami, M.~M. Hassan, P.~Pace, G.~Alai, K.~Lin, and
  G.~Fortino, ``Deep learning and blockchain with edge computing for 5g-enabled
  drone identification and flight mode detection,'' {\em Ieee Network},
  vol.~35, no.~1, pp.~94--100, 2021.

\bibitem{RN45}
I.~Nemer, T.~Sheltami, I.~Ahmad, A.~U. Yasar, and M.~A.~R. Abdeen, ``Rf-based
  uav detection and identification using hierarchical learning approach,'' {\em
  Sensors}, vol.~21, no.~6, p.~23, 2021.

\bibitem{RN260}
H.~Kolamunna, T.~Dahanayaka, J.~Li, S.~Seneviratne, K.~Thilakaratne, A.~Y.
  Zomaya, and A.~Seneviratne, ``Droneprint: Acoustic signatures for open-set
  drone detection and identification with online data,'' {\em Proceedings of
  the ACM on Interactive, Mobile, Wearable and Ubiquitous Technologies},
  vol.~5, no.~1, pp.~1--31, 2021.

\bibitem{RN224}
S.~Song, Y.~Son, and Y.~Kim, ``Flying drone classification based on
  visualization of acoustic signals with deep neural networks,'' in {\em 2020
  International Conference on Information and Communication Technology
  Convergence (ICTC)}, pp.~546--548, IEEE.

\bibitem{RN279}
J.~Garcia-Gomez, M.~Bautista-Duran, R.~Gil-Pita, I.~Mohino-Herranz,
  M.~Aguilar-Ortega, and C.~Clares-Crespo, {\em Cost-constrained Drone Presence
  Detection through Smart Sound Processing}.
\newblock Icpram: Proceedings of the 8th International Conference on Pattern
  Recognition Applications and Methods, Setubal: Scitepress, 2019.

\bibitem{RN223}
M.~Z. Anwar, Z.~Kaleem, and A.~Jamalipour, ``Machine learning inspired
  sound-based amateur drone detection for public safety applications,'' {\em
  Ieee Transactions on Vehicular Technology}, vol.~68, no.~3, pp.~2526--2534,
  2019.

\bibitem{RN48}
B.~W. Yang, E.~T. Matson, A.~H. Smith, J.~E. Dietz, J.~C. Gallagher, and Ieee,
  {\em UAV Detection System with Multiple Acoustic Nodes Using Machine Learning
  Models}.
\newblock 2019 Third Ieee International Conference on Robotic Computing, New
  York: Ieee, 2019.

\bibitem{RN77}
S.~Salman, J.~Mir, M.~T. Farooq, A.~N. Malik, and R.~Haleemdeen, {\em Machine
  Learning Inspired Efficient Audio Drone Detection using Acoustic Features},
  pp.~335--339.
\newblock International Bhurban Conference on Applied Sciences and Technology,
  New York: Ieee, 2021.

\bibitem{RN233}
Y.~J. He, I.~Ahmad, L.~Shi, and K.~Chang, ``Svm-based drone sound recognition
  using the combination of hla and wpt techniques in practical noisy
  environment,'' {\em Ksii Transactions on Internet and Information Systems},
  vol.~13, no.~10, pp.~5078--5094, 2019.

\bibitem{RN433}
G.~Ciaburro and G.~Iannace, ``Improving smart cities safety using sound events
  detection based on deep neural network algorithms,'' {\em Informatics-Basel},
  vol.~7, no.~3, p.~17, 2020.

\bibitem{RN164}
S.~Jeon, J.~W. Shin, Y.~J. Lee, W.~H. Kim, Y.~Kwon, H.~Y. Yang, and Ieee, {\em
  Empirical Study of Drone Sound Detection in Real-Life Environment with Deep
  Neural Networks}, pp.~1858--1862.
\newblock European Signal Processing Conference, New York: Ieee, 2017.

\bibitem{RN262}
S.~Al-Emadi, A.~Al-Ali, and A.~Al-Ali, ``Audio-based drone detection and
  identification using deep learning techniques with dataset enhancement
  through generative adversarial networks,'' {\em Sensors}, vol.~21, no.~15,
  p.~26, 2021.

\bibitem{RN261}
S.~Al-Emadi, A.~Al-Ali, A.~Mohammad, A.~Al-Ali, and Ieee, {\em Audio Based
  Drone Detection and Identification using Deep Learning}, pp.~459--464.
\newblock International Wireless Communications and Mobile Computing
  Conference, New York: Ieee, 2019.

\bibitem{RN200}
M.~Ohlenbusch, A.~Ahrens, C.~Rollwage, J.~Bitzer, and Ieee, {\em Robust Drone
  Detection for Acoustic Monitoring Applications}, pp.~6--10.
\newblock European Signal Processing Conference, New York: Ieee, 2021.

\bibitem{RN239}
Y.~Seo, B.~Jang, S.~Im, and Ieee, {\em Drone Detection Using Convolutional
  Neural Networks with Acoustic STFT Features}.
\newblock 2018 15th Ieee International Conference on Advanced Video and Signal
  Based Surveillance, New York: Ieee, 2018.

\bibitem{RN169}
J.~Kim, C.~Park, J.~Ahn, Y.~Ko, J.~Park, J.~C. Gallagher, and Ieee, {\em
  Real-time UAV Sound Detection and Analysis System}.
\newblock 2017 Ieee Sensors Applications Symposium, New York: Ieee, 2017.

\bibitem{RN238}
N.~Siriphun, S.~Kashihara, D.~Fall, and A.~Khurat, ``Distinguishing drone types
  based on acoustic wave by iot device,'' in {\em 2018 22nd International
  Computer Science and Engineering Conference (ICSEC)}, pp.~1--4, IEEE.

\bibitem{RN244}
S.~Yoo and H.~Oh, ``Analysis of commercial drone sounds and its
  identification,'' in {\em Proceedings of the International Conference on
  Research in Adaptive and Convergent Systems}, pp.~47--52.

\bibitem{RN306}
A.~E. Ananenkov, D.~V. Marin, V.~M. Nuzhdin, V.~V. Rastorguev, and P.~V.
  Sokolov, {\em Possibilities to Observe Small-Size UAVs in the Prospective
  Airfield Radar}.
\newblock International Conference on Transparent Optical Networks-ICTON, New
  York: Ieee, 2018.

\bibitem{RN152}
T.~Al-Nuaim, M.~Alam, and A.~Aldowesh, {\em Low-Cost Implementation of a
  Multiple-Input Multiple-Output Radar Prototype for Drone Detection},
  pp.~183--186.
\newblock ELMAR Proceedings, New York: Ieee, 2019.

\bibitem{RN154}
A.~Aldowesh, T.~BinKhamis, T.~Alnuaim, and A.~Alzogaiby, ``Low power digital
  array radar for drone detection and micro-doppler classification,'' in {\em
  2019 Signal Processing Symposium (SPSympo)}, pp.~203--206, IEEE.

\bibitem{RN311}
Y.~C. Zhao and Y.~Su, ``The extraction of micro-doppler signal with emd
  algorithm for radar-based small uavs' detection,'' {\em Ieee Transactions on
  Instrumentation and Measurement}, vol.~69, no.~3, pp.~929--940, 2020.

\bibitem{RN307}
Y.~C. Zhao and Y.~Su, ``Cyclostationary phase analysis on micro-doppler
  parameters for radar-based small uavs detection,'' {\em Ieee Transactions on
  Instrumentation and Measurement}, vol.~67, no.~9, pp.~2048--2057, 2018.

\bibitem{RN35}
H.~B. Sun, B.~S. Oh, X.~Guo, and Z.~P. Lin, ``Improving the doppler resolution
  of ground-based surveillance radar for drone detection,'' {\em Ieee
  Transactions on Aerospace and Electronic Systems}, vol.~55, no.~6,
  pp.~3667--3673, 2019.

\bibitem{RN95}
A.~D. Huang, P.~Sevigny, H.~Balajit, S.~Rajan, and Ieee, {\em Radar
  Micro-Doppler-based Rotary Drone Detection using Parametric Spectral
  Estimation Methods}.
\newblock IEEE Sensors, New York: Ieee, 2020.

\bibitem{RN46}
J.~F. Ren and X.~D. Jiang, ``A three-step classification framework to handle
  complex data distribution for radar uav detection,'' {\em Pattern
  Recognition}, vol.~111, p.~11, 2021.

\bibitem{RN114}
S.~Zulkifli, A.~Balleri, and Ieee, {\em Design and Development of K-Band FMCW
  Radar for Nano-Drone Detection}.
\newblock IEEE Radar Conference, New York: Ieee, 2020.

\bibitem{RN317}
W.~Y. Zhang and G.~Li, ``Detection of multiple micro-drones via cadence
  velocity diagram analysis,'' {\em Electronics Letters}, vol.~54, no.~7,
  pp.~441--442, 2018.

\bibitem{RN59}
G.~Sacco, E.~Pittella, E.~Piuzzi, S.~Pisa, and Ieee, {\em A MISO radar system
  for drone localization}, pp.~549--553.
\newblock IEEE Metrology for AeroSpace, New York: Ieee, 2018.

\bibitem{RN314}
C.~Ozdemir, ``Radar cross section analysis of unmanned aerial vehicles using
  predics,'' {\em International Journal of Engineering and Geosciences},
  vol.~5, no.~3, pp.~144--149, 2020.

\bibitem{RN119}
V.~Semkin, M.~S. Yin, Y.~Q. Hu, M.~Mezzavilla, S.~Rangan, and Ieee, {\em Drone
  Detection and Classification Based on Radar Cross Section Signatures}.
\newblock 2020 International Symposium on Antennas and Propagation, New York:
  Ieee, 2021.

\bibitem{RN76}
A.~D. de~Quevedo, F.~I. Urzaiz, J.~G. Menoyo, A.~A. Lopez, and Ieee, {\em Drone
  Detection and RCS Measurements with Ubiquitous Radar}.
\newblock 2018 International Conference on Radar, New York: Ieee, 2018.

\bibitem{RN112}
S.~Rzewuski, K.~Kulpa, B.~Salski, P.~Kopyt, K.~Borowiec, M.~Malanowski, and
  P.~Samczyński, ``Drone rcs estimation using simple experimental measurement
  in the wifi bands,'' in {\em 2018 22nd International Microwave and Radar
  Conference (MIKON)}, pp.~695--698, IEEE.

\bibitem{RN298}
M.~Guo, Y.~Lin, Z.~S. Sun, Y.~Q. Fu, and Ieee, {\em Research on Monostatic
  Radar Cross Section Simulation of Small Unmanned Aerial Vehicles}.
\newblock 2018 International Conference on Microwave and Millimeter Wave
  Technology, New York: Ieee, 2018.

\bibitem{RN165}
R.~Nakamura and H.~Hadama, ``Characteristics of ultra-wideband radar echoes
  from a drone,'' {\em Ieice Communications Express}, vol.~6, no.~9,
  pp.~530--534, 2017.

\bibitem{RN105}
R.~Nakamura, H.~Hadama, and A.~Kajiwara, ``Ultra-wideband radar reflectivity of
  a drone in millimeter wave band,'' {\em Ieice Communications Express},
  vol.~7, no.~9, pp.~341--346, 2018.

\bibitem{RN166}
T.~Mizushima, R.~Nakamura, H.~Hadama, and Ieee, {\em Reflection Characteristics
  of Ultra-wideband Radar Echoes from Various Drones in Flight}, pp.~30--33.
\newblock IEEE Topical Conference on Wireless Sensors and Sensor Networks, New
  York: Ieee, 2020.

\bibitem{RN53}
M.~Passafiume, N.~Rojhani, G.~Collodi, and A.~Cidronali, ``Modeling small uav
  micro-doppler signature using millimeter-wave fmcw radar,'' {\em
  Electronics}, vol.~10, no.~6, p.~16, 2021.

\bibitem{RN220}
J.~Ochodnický, Z.~Matousek, M.~Babjak, and J.~Kurty, ``Drone detection by
  ku-band battlefield radar,'' in {\em 2017 International Conference on
  Military Technologies (ICMT)}, pp.~613--616, IEEE.

\bibitem{RN304}
J.~S. Patel, F.~Fioranelli, and D.~Anderson, ``Review of radar classification
  and rcs characterisation techniques for small uavs or drones,'' {\em Iet
  Radar Sonar and Navigation}, vol.~12, no.~9, pp.~911--919, 2018.

\bibitem{RN151}
S.~A. Musa, R.~Abdullah, A.~Sali, A.~Ismail, N.~E.~A. Rashid, I.~P. Ibrahim,
  and A.~A. Salah, ``A review of copter drone detection using radar systems,''
  {\em Def. S\&T Tech. Bull}, vol.~12, no.~1, pp.~16--38, 2019.

\bibitem{RN87}
A.~Coluccia, G.~Parisi, and A.~Fascista, ``Detection and classification of
  multirotor drones in radar sensor networks: A review,'' {\em Sensors},
  vol.~20, no.~15, p.~22, 2020.

\bibitem{RN71}
B.~Taha and A.~Shoufan, ``Machine learning-based drone detection and
  classification: State-of-the-art in research,'' {\em Ieee Access}, vol.~7,
  pp.~138669--138682, 2019.

\bibitem{RN221}
A.~Carrio, S.~Vemprala, A.~Ripoll, S.~Saripalli, and P.~Campoy, {\em Drone
  Detection Using Depth Maps}, pp.~1032--1037.
\newblock IEEE International Conference on Intelligent Robots and Systems, New
  York: Ieee, 2018.

\bibitem{RN194}
J.~Ryu and S.~Kim, {\em Small infrared target detection by data-driven proposal
  and deep learning-based classification}, vol.~10624 of {\em Proceedings of
  SPIE}.
\newblock Bellingham: Spie-Int Soc Optical Engineering, 2018.

\bibitem{redmon2016you}
J.~Redmon, S.~Divvala, R.~Girshick, and A.~Farhadi, ``You only look once:
  Unified, real-time object detection,'' in {\em Proceedings of the IEEE
  conference on computer vision and pattern recognition}, pp.~779--788, 2016.

\bibitem{RN190}
D.~K. Behera, A.~B. Raj, and Ieee, {\em Drone Detection and Classification
  using Deep Learning}.
\newblock Proceedings of the International Conference on Intelligent Computing
  and Control Systems, New York: Ieee, 2020.

\bibitem{he2016deep}
K.~He, X.~Zhang, S.~Ren, and J.~Sun, ``Deep residual learning for image
  recognition,'' in {\em Proceedings of the IEEE conference on computer vision
  and pattern recognition}, pp.~770--778, 2016.

\bibitem{RN32}
X.~D. Zhang and K.~Chandramouli, {\em Critical Infrastructure Security Against
  Drone Attacks Using Visual Analytics}, vol.~11754 of {\em Lecture Notes in
  Computer Science}, pp.~713--722.
\newblock Cham: Springer International Publishing Ag, 2019.

\bibitem{RN159}
J.~Park, D.~H. Kim, Y.~S. Shin, S.~H. Lee, and Ieee, {\em A Comparison of
  Convolutional Object Detectors for Real-time Drone Tracking Using a PTZ
  Camera}, pp.~696--699.
\newblock International Conference on Control Automation and Systems, New York:
  Ieee, 2017.

\bibitem{RN344}
H.~Liu, F.~C. Qu, Y.~J. Liu, W.~Zhao, Y.~T. Chen, and Iop, {\em A drone
  detection with aircraft classification based on a camera array}, vol.~322 of
  {\em IOP Conference Series-Materials Science and Engineering}.
\newblock Bristol: Iop Publishing Ltd, 2018.

\bibitem{RN96}
E.~Unlu, E.~Zenou, N.~Riviere, and P.-E. Dupouy, ``Deep learning-based
  strategies for the detection and tracking of drones using several cameras,''
  {\em IPSJ Transactions on Computer Vision and Applications}, vol.~11, no.~1,
  pp.~1--13, 2019.

\bibitem{RN219}
M.~Wu, W.~Xie, X.~Shi, P.~Shao, and Z.~Shi, ``Real-time drone detection using
  deep learning approach,'' in {\em International Conference on Machine
  Learning and Intelligent Communications}, pp.~22--32, Springer.

\bibitem{RN50}
S.~A. Hassan, T.~Rahim, S.~Y. Shin, and Ieee, {\em Real-time UAV Detection
  based on Deep Learning Network}, pp.~630--632.
\newblock International Conference on Information and Communication Technology
  Convergence, New York: Ieee, 2019.

\bibitem{RN170}
A.~Koksal, K.~G. Ince, and A.~Alatan, ``Effect of annotation errors on drone
  detection with yolov3,'' in {\em Proceedings of the IEEE/CVF Conference on
  Computer Vision and Pattern Recognition Workshops}, pp.~1030--1031.

\bibitem{RN407}
K.~Madasamy, V.~Shanmuganathan, V.~Kandasamy, M.~Y. Lee, and M.~Thangadurai,
  ``Osddy: embedded system-based object surveillance detection system with
  small drone using deep yolo,'' {\em Eurasip Journal on Image and Video
  Processing}, vol.~2021, no.~1, p.~14, 2021.

\bibitem{RN101}
A.~Schumann, L.~Sommer, J.~Klatte, T.~Schuchert, J.~Beyerer, and Ieee, {\em
  Deep Cross-Domain Flying Object Classification for Robust UAV Detection}.
\newblock 2017 14th Ieee International Conference on Advanced Video and Signal
  Based Surveillance, New York: Ieee, 2017.

\bibitem{RN445}
A.~Coluccia, A.~Fascista, A.~Schumann, L.~Sommer, M.~Ghenescu, T.~Piatrik,
  G.~De~Cubber, M.~Nalamati, A.~Kapoor, M.~Saqib, N.~Sharma, M.~Blumenstein,
  V.~Magoulianitis, D.~Ataloglou, A.~Dimou, D.~Zarpalas, P.~Daras, C.~Craye,
  S.~Ardjoune, D.~de~la Iglesia, M.~Mendez, R.~Dosil, I.~Gonzalez, and Ieee,
  {\em Drone-vs-Bird Detection Challenge at IEEE AVSS2019}.
\newblock 2019 16th Ieee International Conference on Advanced Video and Signal
  Based Surveillance, New York: Ieee, 2019.

\bibitem{RN207}
C.~Craye, S.~Ardjoune, and Ieee, {\em Spatio-temporal Semantic Segmentation for
  Drone Detection}.
\newblock 2019 16th Ieee International Conference on Advanced Video and Signal
  Based Surveillance, New York: Ieee, 2019.

\bibitem{RN74}
A.~Coluccia, A.~Fascista, A.~Schumann, L.~Sommer, A.~Dimou, D.~Zarpalas,
  M.~Mendez, D.~de~la Iglesia, I.~Gonzalez, J.~P. Mercier, G.~Gagne, A.~Mitra,
  and S.~Rajashekar, ``Drone vs. bird detection: Deep learning algorithms and
  results from a grand challenge,'' {\em Sensors}, vol.~21, no.~8, p.~27, 2021.

\bibitem{RN25}
L.~Du, C.~Q. Gao, Q.~Feng, C.~Wang, and J.~Liu, {\em Small UAV Detection in
  Videos from a Single Moving Camera}, vol.~773 of {\em Communications in
  Computer and Information Science}, pp.~187--197.
\newblock Singapore: Springer-Verlag Singapore Pte Ltd, 2017.

\bibitem{RN83}
U.~Seidaliyeva, D.~Akhmetov, L.~Ilipbayeva, and E.~T. Matson, ``Real-time and
  accurate drone detection in a video with a static background,'' {\em
  Sensors}, vol.~20, no.~14, p.~19, 2020.

\bibitem{RN9}
A.~Sharjeel, S.~A.~Z. Naqvi, and M.~Ahsan, ``Real time drone detection by
  moving camera using corola and cnn algorithm,'' {\em Journal of the Chinese
  Institute of Engineers}, vol.~44, no.~2, pp.~128--137, 2021.

\bibitem{RN70}
P.~Neduchal, F.~Berka, and M.~Zelezny, {\em Stationary Device for Drone
  Detection in Urban Areas}, vol.~10459 of {\em Lecture Notes in Artificial
  Intelligence}, pp.~162--169.
\newblock Cham: Springer International Publishing Ag, 2017.

\bibitem{RN198}
D.~Lee, W.~G. La, H.~Kim, and Ieee, {\em Drone Detection and Identification
  System using Artificial Intelligence}, pp.~1131--1133.
\newblock International Conference on Information and Communication Technology
  Convergence, New York: Ieee, 2018.

\bibitem{RN43}
P.~A. Prates, R.~Mendonça, A.~Lourenço, F.~Marques, J.~P. Matos-Carvalho, and
  J.~Barata, ``Vision-based uav detection and tracking using motion
  signatures,'' in {\em 2018 IEEE Industrial Cyber-Physical Systems (ICPS)},
  pp.~482--487, IEEE.

\bibitem{RN150}
K.~Abbasi, A.~Batool, M.~A. Asghar, A.~Saeed, M.~J. Khan, and M.~ur~Rehman, ``A
  vision-based amateur drone detection algorithm for public safety
  applications,'' in {\em 2019 UK/China Emerging Technologies (UCET)},
  pp.~1--5, IEEE.

\bibitem{RN241}
A.~Fernandes, M.~Baptista, L.~Fernandes, P.~Chaves, and Ieee, {\em Drone,
  Aircraft and Bird Identification in Video Images Using Object Tracking and
  Residual Neural Networks}.
\newblock International Conference on Electronics Computers and Artificial
  Intelligence, New York: Ieee, 2019.

\bibitem{RN5}
A.~J. Garcia, J.~M. Lee, and D.~S. Kim, ``Anti-drone system: A visual-based
  drone detection using neural networks,'' in {\em 2020 International
  Conference on Information and Communication Technology Convergence (ICTC)},
  pp.~559--561, IEEE.

\bibitem{RN236}
S.~Srigrarom, N.~J.~L. Sie, H.~M. Cheng, K.~H. Chew, M.~Lee, P.~Ratsamee, and
  Ieee, {\em Multi-camera Multi-drone Detection, Tracking and Localization with
  Trajectory-based Re-identification}.
\newblock 2021 Second International Symposium on Instrumentation, Control,
  Artificial Intelligence, and Robotics, New York: Ieee, 2021.

\bibitem{RN44}
Q.~Dong and Q.~H. Zou, {\em Visual UAV detection method with online feature
  classification}.
\newblock Proceedings of 2017 Ieee 2nd Information Technology, Networking,
  Electronic and Automation Control Conference, New York: Ieee, 2017.

\bibitem{RN82}
Z.~Z. Wang, L.~Qi, Y.~Tie, Y.~Ding, Y.~Bai, and Ieee, {\em Drone Detection
  Based On FD-HOG Descriptor}, pp.~433--436.
\newblock International Conference on Cyber-Enabled Distributed Computing and
  Knowledge Discovery, New York: Ieee, 2018.

\bibitem{RN92}
H.~Sun, W.~Geng, J.~Q. Shen, N.~Z. Liu, D.~Liang, and H.~Y. Zhou, ``Deeper ssd:
  Simultaneous up-sampling and down-sampling for drone detection,'' {\em Ksii
  Transactions on Internet and Information Systems}, vol.~14, no.~12,
  pp.~4795--4815, 2020.

\bibitem{RN267}
N.~Shijith, P.~Poornachandran, V.~G. Sujadevi, M.~M. Dharmana, and Ieee, {\em
  Breach Detection and Mitigation of UAVs Using Deep Neural Network}.
\newblock 2017 Recent Developments in Control, Automation and Power
  Engineering, New York: Ieee, 2017.

\bibitem{RN111}
E.~Unlu, E.~Zenou, and N.~Riviere, {\em Generic Fourier Descriptors for
  Autonomous UAV Detection}.
\newblock Proceedings of the 7th International Conference on Pattern
  Recognition Applications and Methods, Setubal: Scitepress, 2018.

\bibitem{RN455}
E.~Unlu, E.~Zenou, and N.~Riviere, ``Using shape descriptors for uav
  detection,'' {\em Electronic Imaging}, vol.~2018, no.~9, pp.~128--1--128--5,
  2018.

\bibitem{RN69}
M.~Nalamati, A.~Kapoor, M.~Saqib, N.~Sharma, M.~Blumenstein, and Ieee, {\em
  Drone Detection in Long-range Surveillance Videos}.
\newblock 2019 16th Ieee International Conference on Advanced Video and Signal
  Based Surveillance, New York: Ieee, 2019.

\bibitem{RN72}
H.~Sun, J.~Yang, J.~Q. Shen, D.~Liang, N.~Z. Liu, and H.~Y. Zhou, ``Tib-net:
  Drone detection network with tiny iterative backbone,'' {\em Ieee Access},
  vol.~8, pp.~130697--130707, 2020.

\bibitem{RN257}
L.~Tao, T.~Hong, Y.~C. Guo, H.~Y. Chen, J.~M. Zhang, and Ieee, {\em Drone
  identification based on CenterNet-TensorRT}.
\newblock IEEE International Symposium on Broadband Multimedia Systems and
  Broadcasting, New York: Ieee, 2020.

\bibitem{RN272}
A.~Hernandez, C.~Copot, R.~De~Keyser, T.~Vlas, and I.~Nascu, {\em
  Identification and Path Following Control of an AR.Drone Quadrotor}.
\newblock 2013 17th International Conference on System Theory, Control and
  Computing, New York: Ieee, 2013.

\bibitem{RN157}
C.~Shinde, R.~Lima, K.~Das, and Ieee, {\em Multi-view Geometry and Deep
  Learning Based Drone Detection and Localization}.
\newblock 2019 Fifth Indian Control Conference, New York: Ieee, 2019.

\bibitem{RN450}
L.~Wang, J.~Ai, L.~Zhang, and Z.~Xing, ``Design of airport obstacle-free zone
  monitoring uav system based on computer vision,'' {\em Sensors}, vol.~20,
  no.~9, p.~2475, 2020.

\bibitem{RN86}
D.~T.~W. Xun, Y.~L. Lim, S.~Srigrarom, and Ieee, {\em Drone detection using
  YOLOv3 with transfer learning on NVIDIA Jetson TX2}.
\newblock 2021 Second International Symposium on Instrumentation, Control,
  Artificial Intelligence, and Robotics, New York: Ieee, 2021.

\bibitem{RN258}
C.~Ruiz, X.~L. Chen, P.~Zhang, and Ieee, {\em Poster Abstract: Hybrid and
  Adaptive Drone Identification through Motion Actuation and Vision Feature
  Matching}.
\newblock 2017 16th Acm/Ieee International Conference on Information Processing
  in Sensor Networks, New York: Ieee, 2017.

\bibitem{RN17}
D.~H. Lee, ``Cnn-based single object detection and tracking in videos and its
  application to drone detection,'' {\em Multimedia Tools and Applications},
  p.~12.

\bibitem{RN418}
M.~Saqib, N.~Sharma, S.~D. Khan, M.~Blumenstein, and Ieee, {\em A Study on
  Detecting Drones Using Deep Convolutional Neural Networks}.
\newblock 2017 14th Ieee International Conference on Advanced Video and Signal
  Based Surveillance, New York: Ieee, 2017.

\bibitem{RN256}
H.~M. Oh, H.~Lee, M.~Y. Kim, and Ieee, {\em Comparing Convolutional Neural
  Network(CNN) models for machine learning-based drone and bird classification
  of anti-drone system}, pp.~87--90.
\newblock International Conference on Control Automation and Systems, New York:
  Ieee, 2019.

\bibitem{RN31}
U.~Seidaliyeva, M.~Alduraibi, L.~Ilipbayeva, and A.~Almagambetov, ``Detection
  of loaded and unloaded uav using deep neural network,'' in {\em 2020 Fourth
  IEEE International Conference on Robotic Computing (IRC)}, pp.~490--494,
  IEEE.

\bibitem{RN91}
R.~Jin, J.~Q. Jiang, Y.~H. Qi, D.~F. Lin, and T.~Song, ``Drone detection and
  pose estimation using relational graph networks,'' {\em Sensors}, vol.~19,
  no.~6, p.~20, 2019.

\bibitem{RN286}
S.~Jamil, Fawad, M.~Rahman, A.~Ullah, S.~Badnava, M.~Forsat, and S.~S.
  Mirjavadi, ``Malicious uav detection using integrated audio and visual
  features for public safety applications,'' {\em Sensors}, vol.~20, no.~14,
  p.~16, 2020.

\bibitem{RN68}
H.~Liu, Z.~Q. Wei, Y.~T. Chen, J.~Pan, L.~Lin, Y.~F. Ren, and Ieee, {\em Drone
  Detection based on An Audio-assisted Camera Array}.
\newblock 2017 Ieee Third International Conference on Multimedia Big Data, New
  York: Ieee, 2017.

\bibitem{RN336}
B.~H. Kim, D.~Khan, C.~Bohak, W.~Choi, H.~J. Lee, and M.~Y. Kim, ``V-rbnn based
  small drone detection in augmented datasets for 3d ladar system,'' {\em
  Sensors}, vol.~18, no.~11, p.~16, 2018.

\bibitem{RN135}
B.~H. Kim, D.~Khan, W.~Choi, and M.~Y. Kim, {\em Real Time Counter-UAV System
  for Long Distance Small Drones using Double Pan-Tilt Scan Laser Radar},
  vol.~11005 of {\em Proceedings of SPIE}.
\newblock Bellingham: Spie-Int Soc Optical Engineering, 2019.

\bibitem{gyongy2020high}
I.~Gyongy, S.~W. Hutchings, A.~Halimi, M.~Tyler, S.~Chan, F.~Zhu,
  S.~McLaughlin, R.~K. Henderson, and J.~Leach, ``High-speed 3d sensing via
  hybrid-mode imaging and guided upsampling,'' {\em Optica}, vol.~7, no.~10,
  pp.~1253--1260, 2020.

\bibitem{mora2021high}
G.~Mora-Mart{\'\i}n, A.~Turpin, A.~Ruget, A.~Halimi, R.~Henderson, J.~Leach,
  and I.~Gyongy, ``High-speed object detection with a single-photon
  time-of-flight image sensor,'' {\em Optics Express}, vol.~29, no.~21,
  pp.~33184--33196, 2021.

\bibitem{hutchings2019reconfigurable}
S.~W. Hutchings, N.~Johnston, I.~Gyongy, T.~Al~Abbas, N.~A. Dutton, M.~Tyler,
  S.~Chan, J.~Leach, and R.~K. Henderson, ``A reconfigurable 3-d-stacked spad
  imager with in-pixel histogramming for flash lidar or high-speed
  time-of-flight imaging,'' {\em IEEE Journal of Solid-State Circuits},
  vol.~54, no.~11, pp.~2947--2956, 2019.

\bibitem{ronneberger2015u}
O.~Ronneberger, P.~Fischer, and T.~Brox, ``U-net: Convolutional networks for
  biomedical image segmentation,'' in {\em International Conference on Medical
  image computing and computer-assisted intervention}, pp.~234--241, Springer,
  2015.

\bibitem{RN11}
Y.~Wang, Y.~Chen, J.~Choi, and C.-C.~J. Kuo, ``Towards visible and thermal
  drone monitoring with convolutional neural networks,'' {\em APSIPA
  Transactions on Signal and Information Processing}, vol.~8, 2019.

\bibitem{RN347}
Y.~R. Chen, P.~Aggarwal, J.~Choi, C.~C.~J. Kuo, and Ieee, {\em A Deep Learning
  Approach to Drone Monitoring}, pp.~686--691.
\newblock Asia-Pacific Signal and Information Processing Association Annual
  Summit and Conference, New York: Ieee, 2017.

\bibitem{RN49}
V.~Magoulianitis, D.~Ataloglou, A.~Dimou, D.~Zarpalas, and P.~Daras, ``Does
  deep super-resolution enhance uav detection?,'' in {\em 2019 16th IEEE
  International Conference on Advanced Video and Signal Based Surveillance
  (AVSS)}, pp.~1--6, IEEE.

\bibitem{RN139}
S.~R. Yellapantula, {\em Synthesizing realistic data for vision based
  drone-to-drone detection}.
\newblock Thesis, 2019.

\end{thebibliography}

%
%
%

\end{document}